\documentclass[lettersize,journal]{IEEEtran}
\usepackage{amsmath,amsfonts}
\usepackage{algorithmic}
\usepackage{algorithm}
\usepackage{array}
\usepackage[caption=false,font=normalsize,labelfont=sf,textfont=sf]{subfig}
\usepackage{textcomp}
\usepackage{stfloats}
\usepackage{url}
\usepackage{color}
\usepackage{verbatim}
\usepackage{graphicx}
\usepackage{cite}
\newcommand{\ie}{\textit{i}.\textit{e}.}
\newcommand{\eg}{\textit{e}.\textit{g}.}
\newcommand{\etal}{\textit{et al}.}

\newcommand{\xblue}{\textcolor{black}}

\begin{document}

\title{Towards High-quality HDR Deghosting with Conditional Diffusion Models}

\author{Qingsen Yan,
        Tao Hu,
        Yuan Sun,
        Hao Tang,
        Yu Zhu,
        Wei Dong,
        Luc Van Gool 
        and~Yanning Zhang
\thanks{Q. Yan, Y. Zhu and Y. Zhang are with the School of Computer Science, Northwestern Polytechnical University.}
\thanks{T. Hu, Y. Sun and W. Dong are with the School of Computer Science, Xi'an University of Architecture and Technology.}
\thanks{H. Tang and L. Van Gool are with the School of Computer Science, ETH Zurich.}
\thanks{Corresponding author: Yanning Zhang}
}




\maketitle

\begin{abstract}
High Dynamic Range (HDR) images can be recovered from several Low Dynamic Range (LDR) images by existing Deep Neural Networks (DNNs) techniques.
Despite the remarkable progress, DNN-based methods still generate ghosting artifacts when LDR images have saturation and large motion, which hinders potential applications in real-world scenarios.
To address this challenge, we formulate the HDR deghosting problem as an image generation that leverages LDR features as the diffusion model's condition, consisting of the feature condition generator and the noise predictor.
Feature condition generator employs attention and Domain Feature Alignment (DFA) layer to transform the intermediate features to avoid ghosting artifacts.
With the learned features as conditions, the noise predictor leverages a stochastic iterative denoising process for diffusion models to generate an HDR image by steering the sampling process.
Furthermore, to mitigate semantic confusion caused by the saturation problem of LDR images, we design a sliding window noise estimator to sample smooth noise in a patch-based manner. In addition, an image space loss is proposed to avoid the color distortion of the estimated HDR results.
We empirically evaluate our model on benchmark datasets for HDR imaging. The results demonstrate that our approach achieves state-of-the-art performances and well generalization to real-world images.
\end{abstract}


\section{Introduction} 
\begin{figure}[tb]
\centering
\includegraphics[width=1\linewidth]{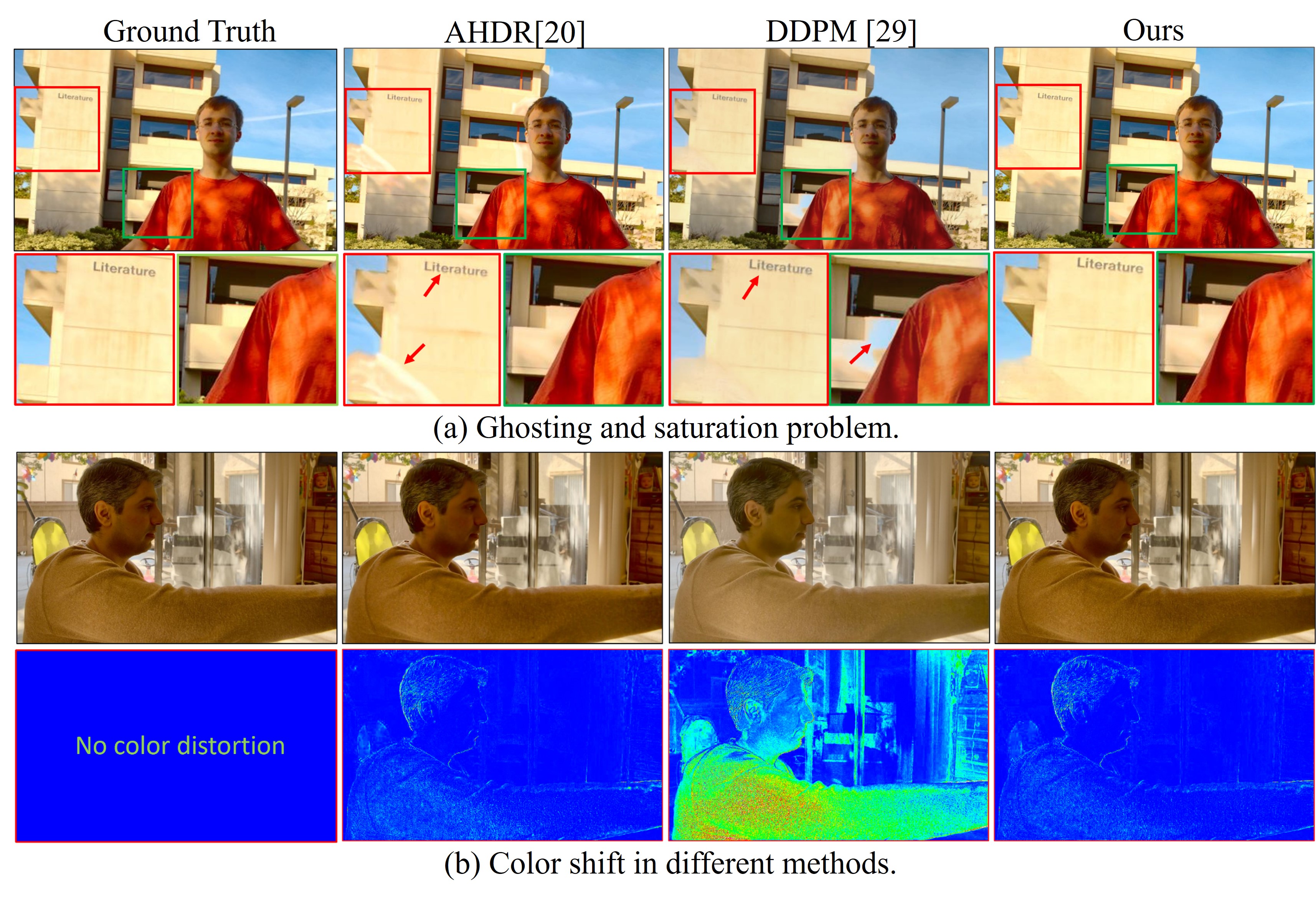}
\caption{The challenges of HDR imaging. 
As shown in (a), DNN-based methods (\eg, AHDR) still tend to generate ghosting and blurring, and DDPM-based methods have a problem of semantic confusion when recovering overexposed regions (green patch).
In addition, (b) the restored results of DDPM-based methods exhibit apparent color shifts when compared with the ground truth.
}
\label{intur}
\vspace{-0.4cm}
\end{figure}
\IEEEPARstart{N}{atural} luminance values have a wide visual dynamic range. However, digital photography sensors typically capture images with limited illumination variation, resulting in low dynamic range (LDR) photos. As a result, LDR images often contain over- or under-exposed regions, which fail to meet human visual expectations for brightness and darkness.
To capture scenes with a broad illumination range, High Dynamic Range (HDR) imaging techniques \cite{Debevec1997Recovering,Picard95onbeing,Reinhard2005High} were developed to generate images covering a wide luminance range; \xblue{since then, these techniques have rapidly advanced and found widespread use in various applications like saliency detection \cite{9056820} and video compression \cite{7095562}.}
Using information from LDR images with varying exposures, multiple exposure fusion \cite{Debevec1997Recovering,Mertens2007Exposure,yan2019enhancing} offers a feasible approach to HDR imaging that endeavors to restore absent details.
Although this method can generate approving HDR images on static scenes, they often result in ghosting artifacts on dynamic scenes caused by object motion or camera shifts, which limits the practical application of HDR imaging in the real world.
Therefore, ghost-free image reconstruction has obtained significant attention from numerous researchers.

Traditional methods often employ motion rejection \cite{Heo2011ACCV,yan2017high,Gallo2009Artifact,Grosch2006}, motion registration \cite{Ward2012,Tomaszewska07,Bogoni2000,Zimmer2011Freehand} and patch matching \cite{Sen2012,Hu2013deghosting,Zheng2013Hybrid} approaches to remove or align the motion regions.
The effectiveness of these methods relies heavily on the performance of preprocessing techniques (\eg, optical flow, motion detection). However, these techniques can be error-prone and less effective when dealing with large-scale foreground motion.
With the success of deep neural network (DNN), numerous DNN-based methods \cite{Kalantari2017Deep,yan2019attention,wu2018deep,yan2020ghost} have been proposed for HDR image reconstruction, which utilizes convolution neural networks (CNNs) or vision transformers (ViT) \cite{liu2022ghost, song2022selective} to generate approving HDR images.
\xblue{In most cases, the necessary information for the overexposed position in the reference frame may not be available in other LDR images when there is object motion or camera movement.
Since DNN-based approaches rely on applying $\mathcal{L}_1$ or $\mathcal{L}_2$ loss to facilitate the network in understanding the intricate relationship between LDRs and ground truth, they cannot produce approving HDR images when motion and saturation are present simultaneously.
This issue can be attributed to these methods' lack of generative capabilities to hallucinate the content.}


\xblue{To address this, deep generative models like GANs \cite{goodfellow2020generative} can generate more realistic details for regions with missing information. \cite{niu2021hdr,9826814} incorporate adversarial losses to produce missing content for HDRs when LDRs have large object motions. Though the adversarial loss can alleviate this issue, these approaches require careful adjustment during training, might overfit certain visual features or data distribution, and might hallucinate new content and artifacts. Recently, Denoising Diffusion Probability Models (DDPM) \cite{pmlr-v37-sohl-dickstein15,NEURIPS2020_4c5bcfec} have shown impressive performance in image synthesis and restoration tasks. DDPM generates high-fidelity images through a stochastic iterative denoising process from pure Gaussian noise. Compared to other generative models such as GANs, DDPM produces a more accurate target distribution without encountering optimization instability or mode collapse.}

Generating vivid images from the input using the Denoising Diffusion Probability Model \cite{pmlr-v37-sohl-dickstein15,NEURIPS2020_4c5bcfec} is a potential solution that outperforms GANs in image-to-image translation tasks, achieving unprecedented performance.
However, producing high-quality HDR images using DDPM is a non-trivial task that involves overcoming several obstacles.
\textit{First}, since the large motion exists in LDR images, concatenating condition (\ie, LDR images) to the noisy image, as was done in SR3 \cite{saharia2022image}, undoubtedly causes ghosting artifacts in HDR images.
\textit{Second}, DDPM tends to use all the information of LDR images as the condition to predict the content of saturated regions, which often causes semantic confusion (\eg, using the sky to fill building) in HDR results (Fig. \ref{intur}(a)).
\textit{Last}, there are noticeable luminance differences in HDR images, the denoising network only learns a probability distribution of the added noise in each step but neglects space constraints on HDR images.
Although this operation can produce images corresponding to human perception, it often has an observable \xblue{color distortion} (Fig. \ref{intur}(b)), underperforming the regression baseline in conventional automated evaluation measures, \eg, \textit{PSNR} and \textit{SSIM}.

\xblue{To alleviate the above challenges and promote the development of diffusion in the HDR image field, we first propose a DDPM-based network to effectively generate a high-quality HDR image in various situations, even for extreme cases (\eg, saturation and occlusion).}
Concretely, to provide a reasonable condition for guiding generation, we employ the attention and Domain Feature Align (DFA) layer to transform the features of the intermediate layer of the network to avoid ghosting artifacts. Note that, compared with \cite{saharia2022image}, which directly uses the image as a condition, the DFA applies affine transformation spatially on feature maps to align the domain features (\ie, the features of differently exposed images).
In addition, to mitigate semantic confusion, we design a sliding window noise estimator to sample smooth global noise in a patch-based manner. Thanks to this operation, our approach can generate better local details while effectively
guaranteeing higher fidelity between locally adjacent blocks. 
Furthermore, we propose the image space loss that preserves color information and provides image pixel-level constraints without modifying the diffusion and sampling process. This effectively eases the \xblue{color distortion} problem of DDPM-based HDR imaging.

The contributions are summarized as follows:
\begin{itemize}
    \item We propose the ﬁrst DDPM-based method for HDR reconstruction from multi-exposed LDR images. By adopting the stochastic iterative denoising process, our method can produce reliable content for HDR images when LDR images contain large object motions.
    \item To generate a better condition, we propose a new structure that utilizes the implicitly aligned LDR features to generate a pair of modulation parameters to apply affine transformation spatially on feature maps of the network.
 \item To avoid semantic confusion, we propose sliding-window noise estimation to learn the global noise and focus on the relationship between adjacent pixels. A pixel space loss is also proposed to performer perception-distortion trade-off and alleviate \xblue{color distortion}.
\end{itemize}

\section{Related Work}
\noindent\textbf{Alignment-based Method.}
These methods try to align the motion regions to reference image and then fuses them to HDR images.
Sen \etal \cite{sen2012robust} formulated the HDR reconstruction as an energy-minimization task that jointly solves patch-based alignment and HDR image reconstruction. Hu \etal \cite{hu2013hdr} proposed a method for image alignment using brightness and gradient information. 
Hafner \etal \cite{hafner2014simultaneous} proposed an energy-minimization method that simultaneously computes the aligned HDR composite and accurate displacement maps.
Li \etal \cite{li2021detail} integrated edge-preserving factors into the fusion method in order to retain the intricate details.
Wang \etal \cite{wang2021unfusion} proposed a fusion network that integrates infrared and visible images using a unified multi-scale densely connected approach.
\xblue{Niu \etal \cite{niu2021hdr} and  Li \etal \cite{9826814} both proposed GAN-based methods that through modifications to the loss, discriminator, and initialization, generated high-quality HDR images.}
Kalantari \etal \cite{Kalantari2017Deep} proposed a representative approach to align low- and high-exposure images to a medium-exposure image \cite{liu2009beyond} and predict HDR image with a DNN.
Yan \etal \cite{yan2019multi} also followed this pipeline. 
Pu \etal \cite{ye2021Progressive} applied deformable convolutions across multi-scale features for pyramidal alignment. 
\xblue{These methods struggle to produce satisfactory HDR images when motion and saturation occur together. While adversarial loss can mitigate this issue, such approaches need careful tuning during training, are prone to overfitting certain visual features or data distribution, and may hallucinate new content and artifacts.}

\noindent\textbf{Detection-based Methods.}
This kind of method assumes that only object motion exists in multi-exposure images.
Khan \etal \cite{khan2006ghost} iteratively removed the moving object using a probability model. Gallo \etal \cite{Gallo2009Artifact} checked the inconsistent patches in the gradient domain to avoid blocking artifacts caused by inaccurate CRF estimation. 
Oh \etal \cite{Oh2015pami} proposed to reject the misaligned moving components as outliers directly.
With the success of DNNs, Yan \etal \cite{yan2019attention} proposed an attention block to detect the misalignment regions and remove useless regions.
Yan \etal \cite{yan2020deep} employed a non-local structure into a U-net to improve the accuracy of HDR image generation.
These methods often generate vivid HDR images, but the results often have ghosting when saturation and motion exist.

\noindent\textbf{Diffusion Models.}
Diffusion generative models \cite{pmlr-v37-sohl-dickstein15,NEURIPS2020_4c5bcfec} had shown impressive results in multiple domains \cite{NEURIPS2020_4c5bcfec,NEURIPS2021_49ad23d1,kingma2021on,10154005}.
Competitive sampling quality could be reached by introducing the reweighted training objective in \cite{NEURIPS2020_4c5bcfec}, which also establishes a close connection to score-based models \cite{song2019generative,song2020score,vincent2011connection}. Following these works, there are many attempts to improve the sampling process in terms of quality and speed. Improved DDPM~\cite{nichol2021glide} chose to additionally predict the variance of the reverse process of a diffusion model. DDIM \cite{DBLP:conf/iclr/SongME21} accelerated the sampling speed by introducing the non-Markovian diffusion process. Meanwhile, the Latent Diffusion Model\cite{rombach2022high} reduced the computational cost by processing the diffusion process in the latent space of the VQ-GAN\cite{esser2021taming}.
However, recovering the color information of saturation regions is challenging to achieve robustly, particularly with large luminance fluctuation.

\section{The Proposed Method}
\begin{figure*}[htbp] \small
\centering
\includegraphics[width=1\textwidth]{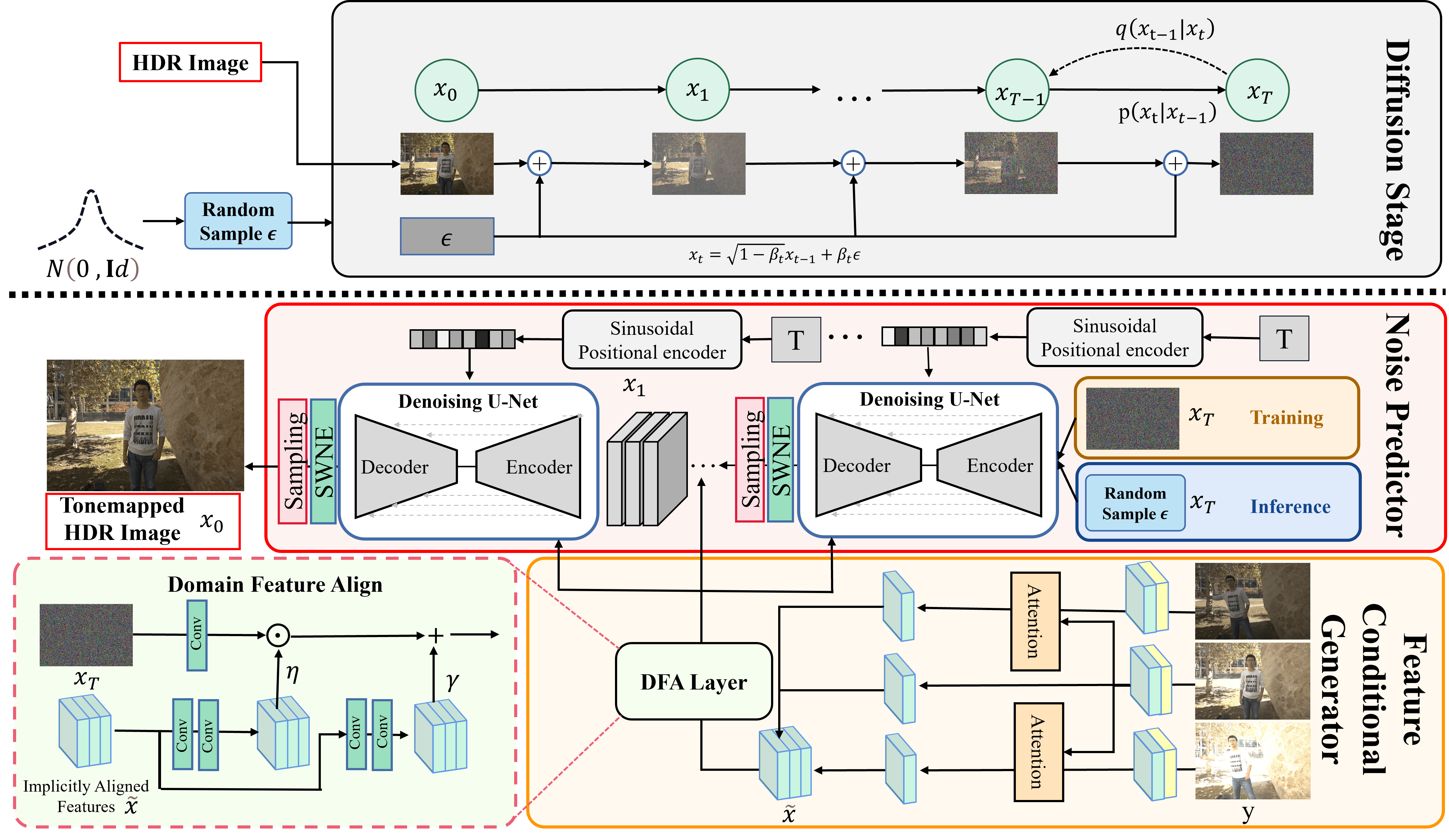}
\caption{The framework of the proposed method. The top figure illustrates the diffusion process, while the bottom figure represents the reverse process, which involves a feature conditional generator and a noise predictor. The feature conditional generator incorporates implicitly aligned LDR features into the noise generator through affine transformation, which guides the model generation.}
\label{frame}
\vspace{-0.4cm}
\end{figure*}

\subsection{Preliminaries}
\label{dpm}
We provide a brief review of ``variance-preserving'' diffusion models from \cite{pmlr-v37-sohl-dickstein15,NEURIPS2020_4c5bcfec}, which gradually transform a Gaussian noise distribution $x_T$ into a complex data distribution $x_0$ using a latent variable model consisting of two processes: the diffusion process and the reverse process.

\noindent\textbf{Diffusion Process.} The diffusion process is a T-step Markov Chain that starts from a clean image distribution $q(x_0)$ and repeatedly injects Gaussian noise $\epsilon$ to the image according to a variance schedule $\beta_1,\cdots,\beta_T$:
\begin{equation}
\xblue{
\begin{aligned}
q(x_t|x_{t-1}) &= \mathcal{N}(x_t;\sqrt{1-\beta_t}x_{t-1},\beta_t\mathbf{I}), \\
q(x_t|x_{0}) &= \mathcal{N}(x_t;\sqrt{\bar{\alpha}_t}x_{0},(1-\bar{\alpha}_t)\mathbf{I}),
\end{aligned}}
\label{diffusion_process}
\end{equation}
where $\alpha_t=1-\beta_t$, $\bar{\alpha}_t=\prod^T_{i=1}\alpha_i$, $\beta_t\in(0,1)$ is a constant hyperparameter that controls the variance of noise added corresponding to the diffusion step $\{1,\cdots, T\}$. The Eq.\ref{diffusion_process} can be further expressed in closed form:
\begin{equation}
  x_t(x_0,\epsilon_t)=\sqrt{\bar{\alpha}_t}x_0+\sqrt{1-\bar{\alpha}_t}\epsilon_t,\epsilon\in(0,\mathbf{I}),
\label{x0xt}
\end{equation}
where the latent variables $x_{1:T}$ have identical shapes to the original image $x_0$.

\noindent\textbf{Reverse Process.} While the diffusion process may seem arbitrary, the reverse process reverses this predeﬁned arbitrary forward process (Eq.~\eqref{diffusion_process}) by the same functional form, which is also deﬁned by a Markov chain \cite{feller2015theory}. The reverse process $q(x_{t-1}|x_t,x_0)$ begins with a standard normal prior $x_T=\mathcal{N}(x_T;\mathbf{0},\mathbf{I})$:
\begin{equation}
\begin{split}
q(x_{t-1}|x_{t},x_0) = \mathcal{N}(x_{t-1};\Tilde{\mu}_t(x_t,x_0),{\Tilde{\beta}_t}\mathbf{I}),
\end{split}
\label{reverse_process}
\end{equation}
where $\Tilde{\mu}_t$ and $\Tilde{\beta}_t$ are the distribution parameters.
The critical component of reverse process is the denoiser network $f_\theta$, which allows us to sample $x_{t-1}$ by using the estimate $p_\theta(x_{t-1}|x_t,t)$ in place of $q(x_{t-1}|x_t,x_0)$:
\begin{align}
\begin{split}
p_\theta(x_{t-1}|x_t,t) &= \mathcal{N}(x_{t-1};\mu_{\theta}(x_t,t),\Sigma_{\theta}(x_t,t)) \\
&= q(x_{t-1}|f_{\theta}(x_t,t)),
\end{split}
\label{reverse_process1}
\end{align}
where the reverse process is parameterized by a neural network $f_{\theta}$ that estimates $\mu_{\theta}(x_t, t)$ and $\Sigma_{\theta}(x_t, t)$. In practice, we consider ﬁxed reverse process variances ($\Sigma_{\theta}(x_t,t)=\Tilde{\beta}_t\mathbf{I}$) and use an alternative parametrization of $\mu_{\theta}$ \cite{ho2020denoising}. We get a simpliﬁed objective $\mathcal{L}_{sim}(\theta)$:
\xblue{
\begin{align}
    \mathbb{E}_{t,x_0,\epsilon_t}[\Vert\epsilon_t-f_{\theta}(\sqrt{{\bar{\alpha}_t}}x_0+\sqrt{1-\bar{\alpha}_t}\epsilon_t,t)\Vert^2].
\end{align}
}

\begin{algorithm}[t]
\caption{Training}
\renewcommand{\algorithmicrequire}{ \textbf{Input:}}
\label{alg:A}  
\begin{algorithmic}[1]
\REQUIRE~~\\  \xblue{LDR-HDR image pairs $(x,y)$}, \\Total diffusion step $T$, Noise schedule $\alpha_{1:T}$.
\STATE \textbf{Initialize:} random initialized feature condition generator $g_\theta$ and noise predictor$f_\theta$.
\REPEAT
\STATE $t \sim Uniform\{1,\cdots,T\}$
\STATE $\epsilon_t\sim \mathcal{N}(0,\mathbf{I})$
\STATE $x_0=\mathcal{T}(x)$ ~~~~~// tonemapping operate as \cite{yan2019attention}
\STATE Take gradient descent step on \\
$\mathbb{E}_{t,x_0,\epsilon_t}[\Vert\epsilon_t-f_{\theta}(\sqrt{{\bar{\alpha}_t}}x_0+\sqrt{1-\bar{\alpha}_t}\epsilon_t,t,g_{\theta}{(y)})\Vert^2]$\\
$\mathbb{E}_{t,x_0,\epsilon_t}[\Vert x_0-\frac{x_t-\sqrt{1-\bar{\alpha}_t}\epsilon_t}{\sqrt{\bar{\alpha}_t}})\Vert^2]$

\UNTIL converged
\end{algorithmic}  
\end{algorithm}

\subsection{HDR Diffusion Model}

Given a dataset of input-output pairs, each pair consists of three LDR images $y$ and an HDR image $x$ (see Fig. \ref{frame}). We aim to learn a parametric approximation to $p(x|y)$ through a stochastic iterative reﬁnement process (Sec.~\ref{dpm}) that maps three LDR images $y$ to a target HDR image $x\in{\mathbb{R}^d}$.
We propose a simple but effective scheme, which learns a conditional reverse process without modifying the diffusion process Eq.~\eqref{diffusion_process}. As shown in Fig.~\ref{frame}, we integrate the feature condition generator branch $g_{\theta}$ into the noise predictor $f_{\theta}$, which guides the recovery of specific HDR images by adding 
LDR information at each reverse time step Eq.~\eqref{reverse_process1}. 
Letting $g_{\theta}$ denotes feature condition generator, the new objective becomes: $\mathcal{L}_{noise}(\theta) = $
\begin{align}
    \mathbb{E}_{t,x_0,\epsilon_t}[\Vert\epsilon_t-f_{\theta}(\sqrt{{\bar{\alpha}_t}}x_0+\sqrt{1-\bar{\alpha}_t}\epsilon_t,t,g_{\theta}{(y)})\Vert^2],
\end{align}
where $g_{\theta}$ does not require an extra loss because the gradient from the loss ﬂows through $f_{\theta}$ into $g_{\theta}$. We include a pseudocode for the modified training procedure in Alg. \ref{alg:A}.

\noindent\textbf{Feature Condition Generator (FCG).} 
Although it is an intuitive solution to guide generation that concatenates three LDR images to the noisy image $x_t$, this approach would undoubtedly affect the quality of image generation (Sec.~\ref{ablation sec}) due to luminance differences and unaligned LDR images. Based on the above problem, as shown in Fig. \ref{frame}, a Domain Feature Align (DFA) layer learns a mapping function $\mathcal{M}$ that outputs a modulation parameter pair $(\eta,~\gamma)$ based on the LDR condition $\Tilde{x}$. The learned parameter pair adaptively influences the outputs by applying an affine transformation spatially to the intermediate feature map in the noise predictor network. 
More precisely, we first generate implicitly aligned features $\Tilde{x}$ employing the attention module (AM) from AHDR \cite{yan2019attention}. 
\xblue{It is worth noting that since our goal is to explore the potential of diffusion models for HDR reconstruction, we adopt a typical attention mechanism-based implicit alignment approach \cite{yan2019attention,liu2022ghost} without more tailored design of the relevant modules.}
The prior $\Tilde{x}$ is modeled by a pair of affine transformation parameters $(\eta,~\gamma)$ through a mapping function $\mathcal{M}: \Tilde{x} \mapsto (\eta,~\gamma)$. In this study, $\mathcal{M}$ is implemented by two consecutive convolutions. Consequently,
\begin{align}
    \Tilde{x}= AM(y),~(\eta,~\gamma)=\mathcal{M}(\Tilde{x}).
\end{align}
The transformation is executed by scaling and shifting the feature maps of the intermediate layer from noise predictor, after obtaining $(\eta,~\gamma)$ through implicitly aligned features $\Tilde{x}$:
\begin{align}
    DFA(F|\eta,~\gamma)=\eta\odot F+\gamma,
\end{align}
where $F$ is obtained from input $x_t$ using a $1\times 1$ convolution, and the $\odot$ symbol denotes element-wise multiplication. As the spatial dimensions of $F$ are preserved, the DFA layer can perform both feature-wise and spatial-wise transformations, allowing for greater flexibility in embedding the implicitly aligned features.

\begin{figure*}[htbp]
\centering
\includegraphics[width=\textwidth]{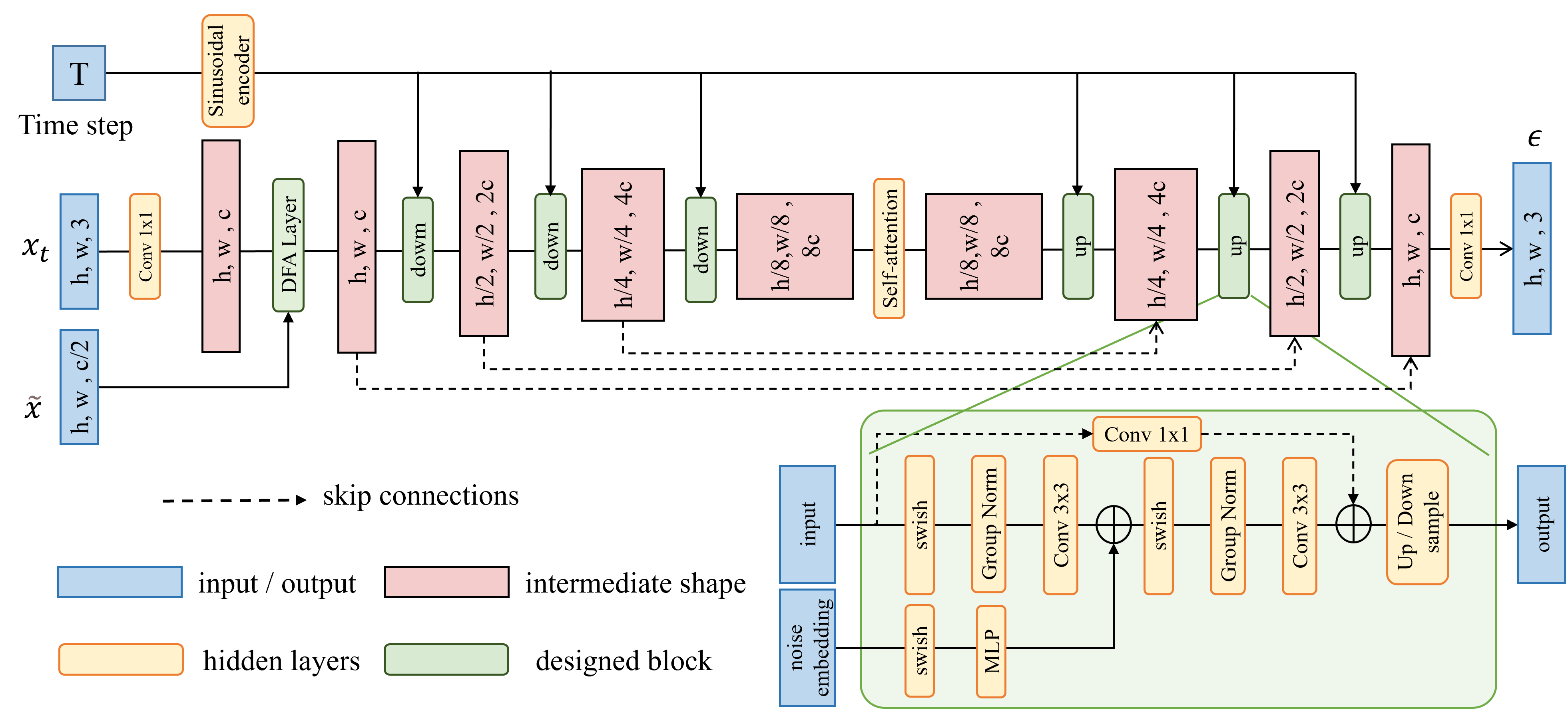}
\caption{The diagram illustrates the U-Net architecture used for the Noise predictor, incorporating $\Tilde{x}$ which represents the implicitly aligned LDR features generated by Attention Network from \cite{yan2019attention}. To simplify comprehension, we display the network architecture with channel multipliers of $\{1,2,4,8\}$. The dimensions W, H, and C correspond to the width, height, and number of channels of the features respectively. }
\label{unet}
\end{figure*}

\noindent\textbf{Noise Predictor.}
The noise generator predicts noise at each timestep of the reverse process, conditioned on the implicitly aligned feature. The network architecture is a modified version of the UNet found in \cite{NEURIPS2020_4c5bcfec}. We replace the original DDPM residual blocks with residual blocks from WideResNet \cite{DBLP:conf/interspeech/HuZLJWKZJ22}, which includes group normalization \cite{wu2018group} and self-attention \cite{vaswani2017attention} blocks in the middle step. To enable parameter sharing across time, we transform the timestep through sinusoidal positional encoding and fuse these embeddings to each residual block using an MLP. Specifically, we first transfer the 3-channel input $x_t$ to the hidden feature $F$ with 128 channels, through a 2D-Convolution. Then, the implicitly aligned LDR features are fused with the DFA layer to provide a better feature condition and fed into the modified UNet. The detailed architecture can be found in the Fig. \ref{unet}.

\subsection{Sliding-window Noise Estimation}
\begin{figure}[tb]
\centering
\includegraphics[width=1\linewidth]{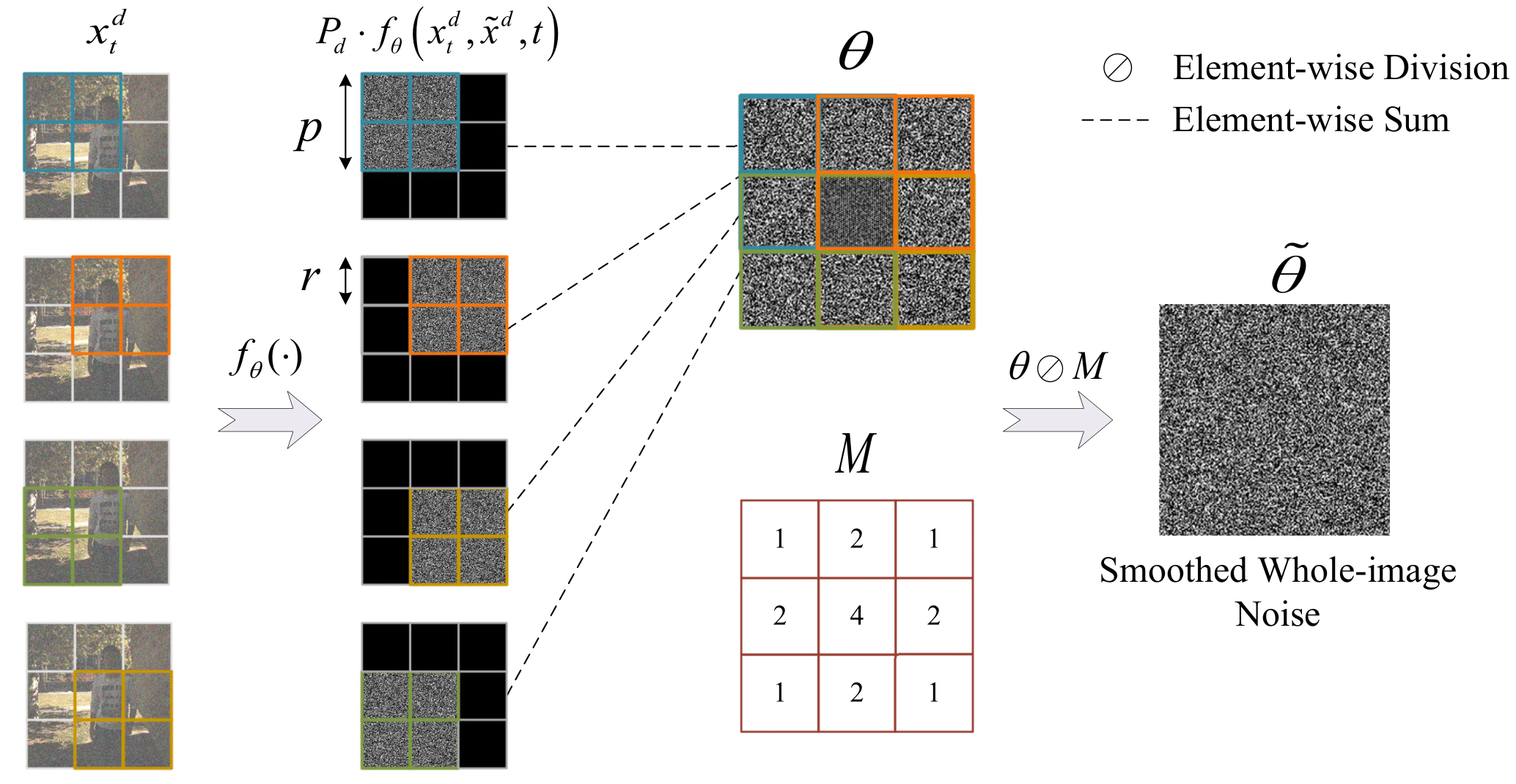}
\caption{Sliding-window Noise Estimation. To facilitate comprehension, we present an intuitive description of the sliding-window noise estimation method. Please refer to lines 6-11 in Algorithm. \ref{alg:B} for the specific steps of the procedure.}
\vspace{-0.4cm}
\label{sw}
\end{figure}

To encourage the model to fully utilize the learned patch statistics and focus on local contextual information, we carefully design a patch-based sampling method called Sliding-window Noise Estimation. 
The typical patch-based approach infers the decomposed image patches and then optimally merges the results, ignoring the intermediate sampling step, and may contain artifacts from independently restored results.

Unlike overlapping the final reconstruction patch after sampling in DDPM, we use the sliding window method (see Fig.~\ref{sw}) to estimate the smoothed whole-image noise in each reverse step. Specifically, we first decompose the input $x_t {\in} \mathbb{R}^{H\times W \times C}$ of arbitrary size by a grid-like arrangement, where each grid cell contains $r\times r$ pixels. Then, we create a sliding window $P_d$ moving over this grid with the step of each grid cell, where the window size is $p{\times} p $ pixels $(p{>}r)$ and $d$ represents the current number of slides. In practice (Alg.~\ref{alg:B}), $P_d$ is a binary mask matrix of the same dimensionality as $x_t$, and $Slide(\cdot)$ denotes extract patch from current location $d$. We define two matrices $\Theta{\in} \mathbb{R}^{H\times W \times C}$ and $M{\in} \mathbb{R}^{H\times W \times C}$ and all elements are 0, where $\Theta$ records the cumulative noise estimation that each pixel receives through the sliding window, and $M$ records the number of times it has received an estimation. Next, we can calculate the mean noise estimate $\Tilde{\Theta}$ of the whole image after $P_d$ passes through the entire image and use the smoothed noise estimation for sampling in each reverse time step. Thanks to these operations, our method can effectively mitigate damage to the learned local posterior, ensuring higher fidelity between locally adjacent blocks and avoiding semantic confusion problems in HDR images.

\begin{algorithm}[t]  \small
\caption{Inference by Sliding-window Noise Estimation}  
\label{alg:B}  
\begin{algorithmic}[1]
\REQUIRE~~\\ $f_\theta$: Noise predictor,\\ $g_\theta$: Feature condition generator,\\
$y$: Input LDR images, $\alpha_{1:T}$: Noise schedule.
\STATE {$x_T\sim\mathcal{N}(\mathbf{0},\mathbf{I})$ } 
\STATE Encoder LDR images $y$ as $\Tilde{x} = g_\theta(y)$ 
\FOR{$t=T,\cdots,1$}
\STATE $z \sim\mathcal{N}(\mathbf{0},\mathbf{I})$, if $t\textgreater 1$, else $z=0$
\STATE Initialize $\Theta = 0$ and $M=0$
\FOR{$d=1,\cdots,D$}
\STATE $x_t^{d}=Slide(P_d \circ x_t)$ and $\Tilde{x}^{d}=Slide(P_d \circ \Tilde{x})$
\STATE $\Theta = \Theta + P_d \cdot f_{\theta}(x_t^{d},\Tilde{x}^{d},t)$
\STATE $M=M+P_d$
\ENDFOR
\STATE $\Tilde{\Theta} = \Theta\oslash M$  ~~~~~//element-wise division
\STATE $x_{t-1}=\sqrt{\bar{\alpha}_{t-1}}(\frac{x_t-\sqrt{1-\bar{\alpha}_{t}}\cdot\Tilde{\Theta}}{\sqrt{\bar{\alpha}_t}}) + \sqrt{1-{\bar{\alpha}_{t-1}}} \cdot\Tilde{\Theta})$
\ENDFOR  
\RETURN $R(x_0)$ ~~~~~// reverse operation of $\mathcal{T}(x)$
\end{algorithmic}  
\end{algorithm}

\subsection{Image-space Loss}
In HDR imaging tasks, the color information of the image is equally important to the perceptual quality of the generated content. Therefore, we propose a novel yet simple approach to providing image space constraints for diffusion models, avoiding the color distortion caused by the luminance difference of LDR images. 
Since the diffusion models aim to fit the probability distribution of noise added during the diffusion process in training, rather than the pixel-level image distribution, traditional methods \cite{saharia2022image} require T-step iterative sampling to obtain the final sampled image result. However, this approach is inefficient and computationally expensive. 
Thus, we attempt to transform the predicted probability distribution into an image distribution using the input noise image $x_t$ and constrain it in image space using $\mathcal{L}_1$ or $\mathcal{L}_2$ loss functions.
This is mathematically feasible, as we have derived the formula by reverse engineering Eq. \eqref{x0xt}:
\begin{equation}
  x_{0,t}=\frac{x_t-\sqrt{1-\bar{\alpha}_t}\epsilon_t}{\sqrt{\bar{\alpha}_t}},
\end{equation}
where $x_{0,t}$ represents the converted image result at $t$ step. Thus, the image space loss function $\mathcal{L}_{image}$ can be written as:
\begin{align}
    \mathbb{E}_{t,x_0,\epsilon_t}[\Vert x_0-\frac{x_t-\sqrt{1-\bar{\alpha}_t}\epsilon_t}{\sqrt{\bar{\alpha}_t}})\Vert^2].
    \label{lour}
\end{align}
It is worth noting that $x_t$ is the input of the network, and $\bar{\alpha}_t$ is a hyperparameter pre-set in the network. When calculating gradients, it can be treated as a constant, so it does not affect the propagation of gradient flow. The final loss $\mathcal{L}_{our}$ can be formulated as:
\begin{align}
    \mathcal{L}_{our} = \mathcal{L}_{noise}+\mathcal{L}_{image}.
\end{align}

\begin{table*}[t] \small
\caption{Evaluation results on Kalantari's \cite{Kalantari2017Deep}.
The best and second best results are highlighted in \textbf{Bold} and \underline{Underline}.}
\label{compare}
	\resizebox{1\linewidth}{!}{%
\begin{tabular}{c|ccccccccccc}
\hline
Models & Hu \cite{Tursun2016data}     & Kalantari \cite{Kalantari2017Deep} & DeepHDR \cite{Wu_2018_ECCV} & AHDR \cite{yan2019attention} & NHDRR \cite{yan2020deep}  & HDRGAN \cite{niu2021hdr} & ADNet\cite{liu2021adnet}  & APNT\cite{chen2022attention}   & ST-HDR \cite{song2022selective} & CA-VIT \cite{liu2022ghost}          & Our             \\ \hline
PSNR-$\mu$~$\uparrow$       & 32.19  & 42.74     & 41.64   & 43.62   & 42.41  & 43.92  & 43.76  & 43.94  & \underline{44.09}  & 43.70  & \textbf{44.11 }   \\
PSNR-L$\uparrow$       & 30.84  & 41.22     & 40.91   & 41.03   & 41.08  & 41.57  & 41.27  & 41.61  & 41.70   & \underline{41.72}  & \textbf{41.73}     \\
SSIM-$\mu$~$\uparrow$       & 0.9716 & 0.9877    & 0.9869  & 0.9900    & 0.9887 & 0.9905 & 0.9904 & 0.9898 & \underline{0.9909} & \textbf{0.9911} & \textbf{0.9911}    \\
SSIM-L~$\uparrow$       & 0.9506 & 0.9848    & 0.9858  & 0.9862  & 0.9861 & 0.9865 & 0.986  & 0.9879 & 0.9872 & \underline{0.9877}    & \textbf{0.9885} \\
HDR-VDP-2~$\uparrow$       & 55.25  & 60.51     & 60.50    & 62.30    & 61.21  & 65.45  & 62.61  & 64.05  & 63.37  & \underline{65.49}           & \textbf{65.52}                \\
LPIPS~$\downarrow$       & 0.0302  & 0.0341     & -    & 0.0166    & -  & 0.0159  & 0.0169  & -  & -  & \underline{0.0111}      &   \textbf{0.0109}           \\ 
FID~$\downarrow$       & 37.27  & 33.30     & -    & 9.43    & -  & 9.32  & 12.42  & -  & -  & \underline{6.21}      &   \textbf{6.20}           \\ 
\hline
\end{tabular}
}
\vspace{-0.4cm}
\end{table*}

\subsection{Implicit Sampling}
Considering the typical diffusion model inference requires relatively large steps (\eg., 1000) to achieve optimal image synthesis quality, \cite{DBLP:conf/iclr/SongME21} accelerated this process by “skipping” steps with appropriate update rules. Since our method only revised the estimation of $\epsilon_{\theta}$, it does not affect the ability to perform implicit sampling, where the sampling method is a non-Markovian forward process:
\begin{align}
\begin{aligned}
q(x_t|x_{t-1},x_0) &= \mathcal{N}(x_{t-1};\Tilde{\mu}_t(x_t,x_0),\sigma_t^2 \mathbf{I}), \\
\Tilde{\mu}_t &= \sqrt{\bar{\alpha}_{t-1}}x_0 + \sqrt{1-{\bar{\alpha}_{t-1}}-\sigma_t^2}\epsilon_t,
\end{aligned}
\end{align}
and implicit sampling using a noise predictor can be executed by:
\begin{align}
\begin{aligned}
   x_{t-1}& = \sqrt{\bar{\alpha}_{t-1}}(\frac{x_t-\sqrt{1-\bar{\alpha}_{t}}\cdot f_{\theta}(x_t,t,g_\theta(y))}{\sqrt{\bar{\alpha}_t}}) \\ &+ \sqrt{1-{\bar{\alpha}_{t-1}}} \cdot f_{\theta}(x_t,t,g_\theta(y)).
\end{aligned}
\end{align}
It is worth noting that this sampling method is independent of the training process. When incorporating Eq.~\eqref{x0xt} into $\Tilde{\mu}_t(x_t,x_0)$, the sampling process will be consistent with the original DPM by setting $\sigma_t^2 = \Tilde{\beta}_t$. Therefore, we can choose a sub-sequence $\{1,\cdots, T^{\prime}\}$ of $\{ 1,\cdots, T\}$ used in training without modifying the training settings, which allows us to generate high-quality HDR images in fewer iterative steps \xblue{(Alg.~\ref{alg:B} line 12)}.

\subsection{Train and Inference}
\xblue{Training models on the tonemapped domain is often essential for existing deep learning-based methods \cite{Kalantari2017Deep,yan2019attention,liu2022ghost,song2022selective,niu2021hdr,9826814,chen2022attention,liu2021adnet,yan2020deep}.
Since traditional DDPM-based methods only learn a probability distribution of the added noise in each step, we devised a ddpm-specific approach that applies the $\mu$-law function to transform the HDR image before adding noise, in order to indirectly achieve a similar effect.}

\textbf{Training.} During the training phase, the input LDR-HDR image pairs $(X, Y)$ in the training set are used to train our model with the total diffusion step $T$ (see Algorithm. \ref{alg:A}).
Since the HDR images are usually displayed after tone mapping, training the network on the tone-mapped images is more effective
than training directly in the HDR domain \cite{Kalantari2017Deep}. \xblue{Given an HDR image  $x$ in the HDR domain, we maps HDR pixel values to a more uniform distribution using $\mu~law$:}
\begin{align}
\begin{split}
x_0=\mathcal{T}(x) &= \frac{log(1+\mu x)}{log(1+\mu)},
\end{split}
\label{tonemaploss}
\end{align}
\xblue{where $\mathcal{T}(\cdot)$ denotes the tone mapping operate, $\mu$ is a parameter deﬁning the amount of compression.} In this work, we always keep $x$ in the range [0, 1], $\mu = 5000$, and train the network with $x_0$ in the tone-mapping case (Alg. \ref{alg:A} line 5).
Before feeding the LDR images to the Attention Network, we first map the input LDR images $Y_i$ to the HDR domain relying on gamma correction to generate a corresponding set of $H_i$. As suggested in \cite{Kalantari2017Deep}, we concatenate images $Y_i$ and $H_i$ along the channel dimension to obtain the 6-channel tensors $y_i = [Y_i,H_i],~i = 1, 2, 3$ as the input of the network. 

\textbf{Inference.} A T-step inference process takes three LDR images (the 6-channel tensors) as input, as illustrated in Algorithm \ref{alg:B}. Unlike the training phase, we encode implicitly aligned LDR information $\Tilde{x}$ by the Attention Network $g_{\theta}$ only once, which speeds up the inference. We random sample a latent variable $x_T \in \mathcal{N}(0,\mathbf{I})$ and convert it to a particular tone-mapped HDR image by a stochastic iterative refinement process using Sliding-window Noise Estimation. Since the training phase is based on a tone-mapping case, we utilize $R(\cdot)$ to restore the final output, which can be expressed as:
\begin{align}
R(x_0) = \frac{e^{x_0 \cdot log(1+\mu)}-1}{\mu}
\end{align}
where $R(\cdot)$ is the reverse process of $\mathcal{T}(\cdot)$. \xblue{We will discuss the motivation for optimizing the network in the tonemapped domain in Sec. \ref{tonemap diss}.}

\section{Experiments}
\begin{table}[h]
\caption{Noise predictor network configurations and parameter choices.}
\centering
\begin{tabular}{cc}
\hline
                               & Hyper-parameters      \\ \hline
Diffusion steps ($T$)            & 1000                  \\
Noise schedule ($\beta_t$)            & linear: 0.0001 → 0.02 \\
Base channels                  & 128                   \\
Channel multipliers            & \{1, 1, 2, 2, 4, 4\}  \\
Residual blocks per resolution & 1                     \\
Time step embedding length     & 512                   \\
Number of parameters           & 74.99$M$                 \\ \hline
\end{tabular}
\label{npconfig}
\end{table}

\begin{figure*}[h]
\centering
\includegraphics[width=1\linewidth]{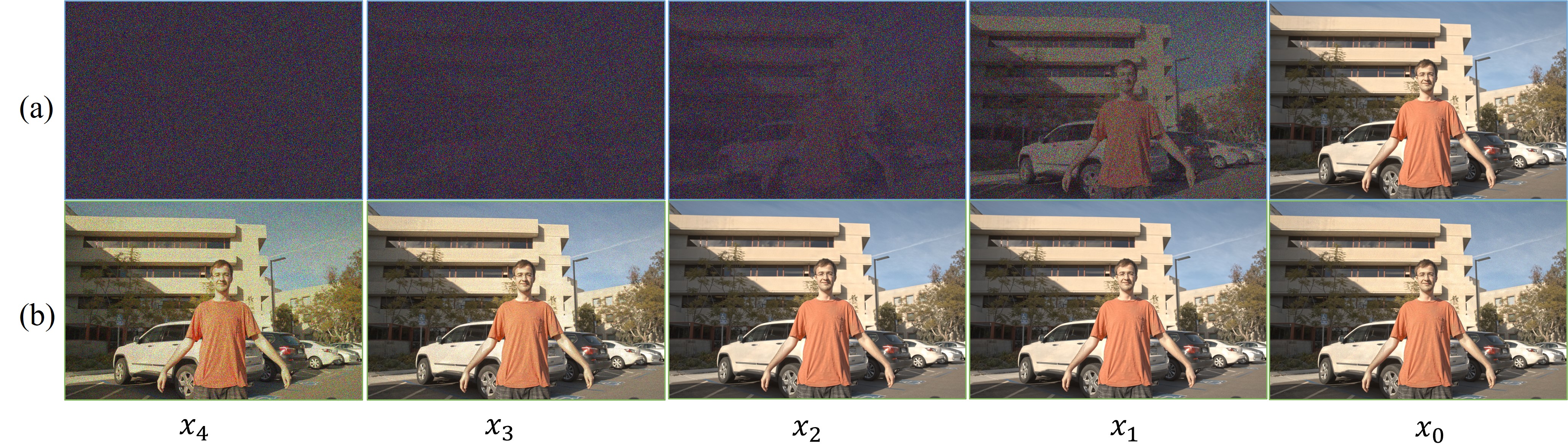}
\caption{Intermediate step denoising results on the testing data from \cite{Kalantari2017Deep}. (a) Results from the intermediate step of directly predicting $x_0$ from $x_t$.
(b) Results from the intermediate step of predicting $x_{t-1}$ from $x_t$ in the intermediate steps.}
\label{inter result1}
\end{figure*}

\begin{figure*}[h]
\centering
\includegraphics[width=0.9\linewidth]{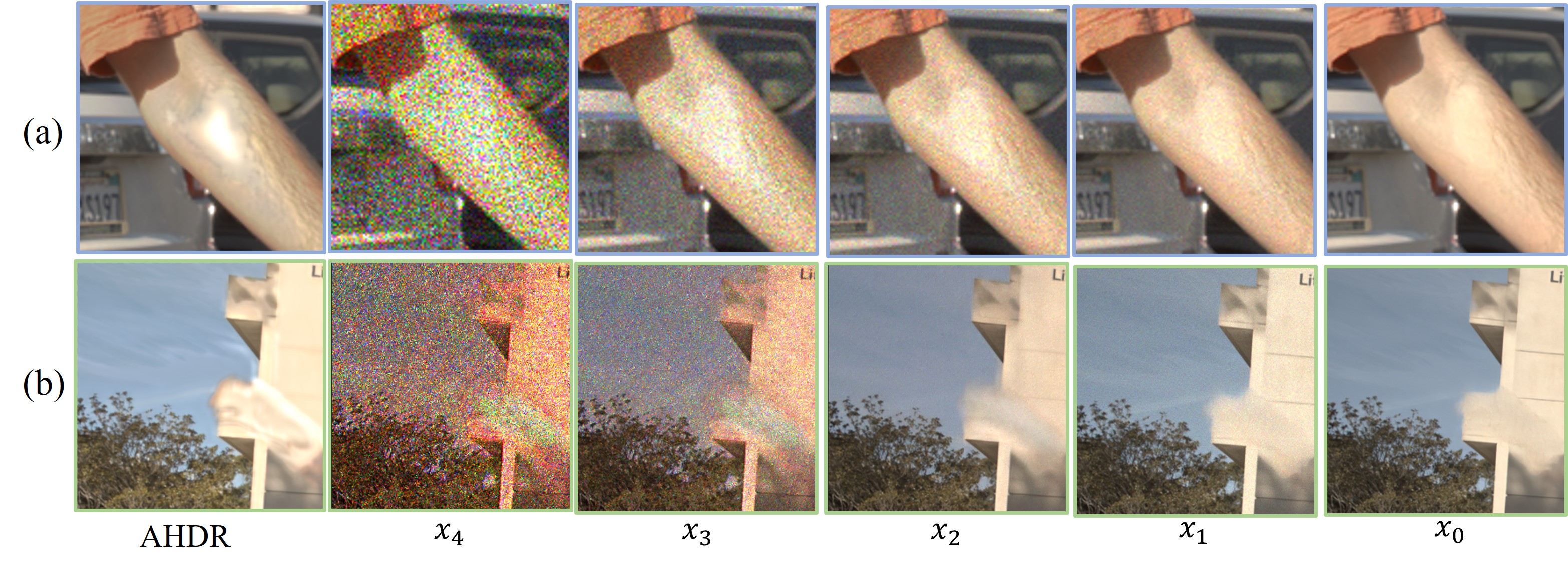}
\caption{Denoising results from the intermediate step of directly predicting $x_0$ from $x_t$ on typical testing data patches from \cite{Kalantari2017Deep}. As the iterative steps proceed, the ghosting artifacts are progressively reduced and more perceptually plausible contents are generated. This demonstrates the effectiveness of DDPM for ghost removal in HDR imaging.}
\label{inter result2}
\end{figure*}

\begin{figure*}[]
\centering
\includegraphics[width=\textwidth]{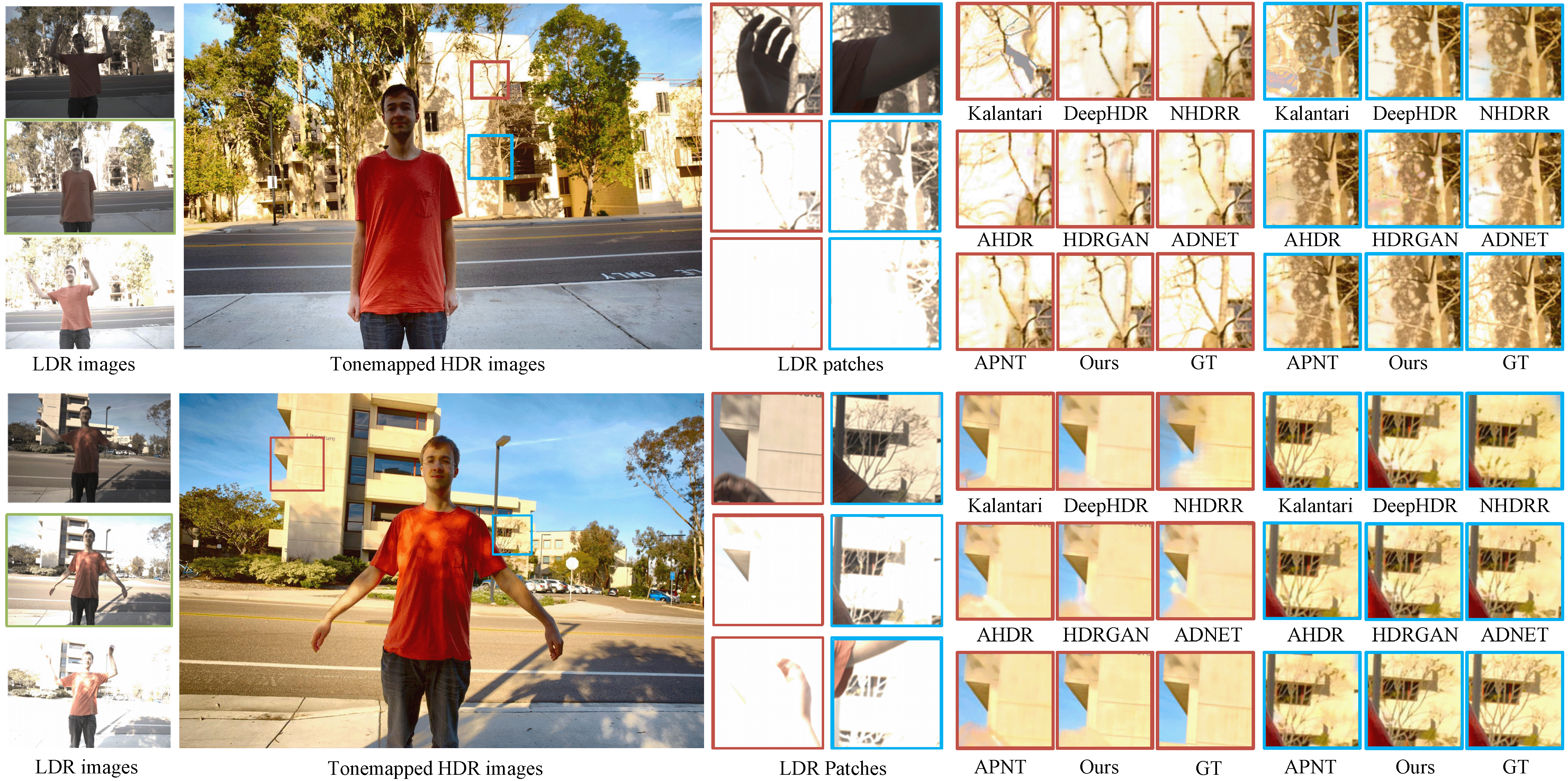}
\caption{Visual comparisons on the testing data from Kalantari \etal\cite{Kalantari2017Deep}. We compare the zoomed-in local areas of the HDR images estimated by our methods and the compared methods. Our model is capable of generating HDR images of exceptional quality, with results that closely align with how humans perceive visual information.}
\label{compare3}
\vspace{-0.4cm}
\end{figure*}

\begin{figure*}[ht]
\centering
\includegraphics[width=\textwidth]{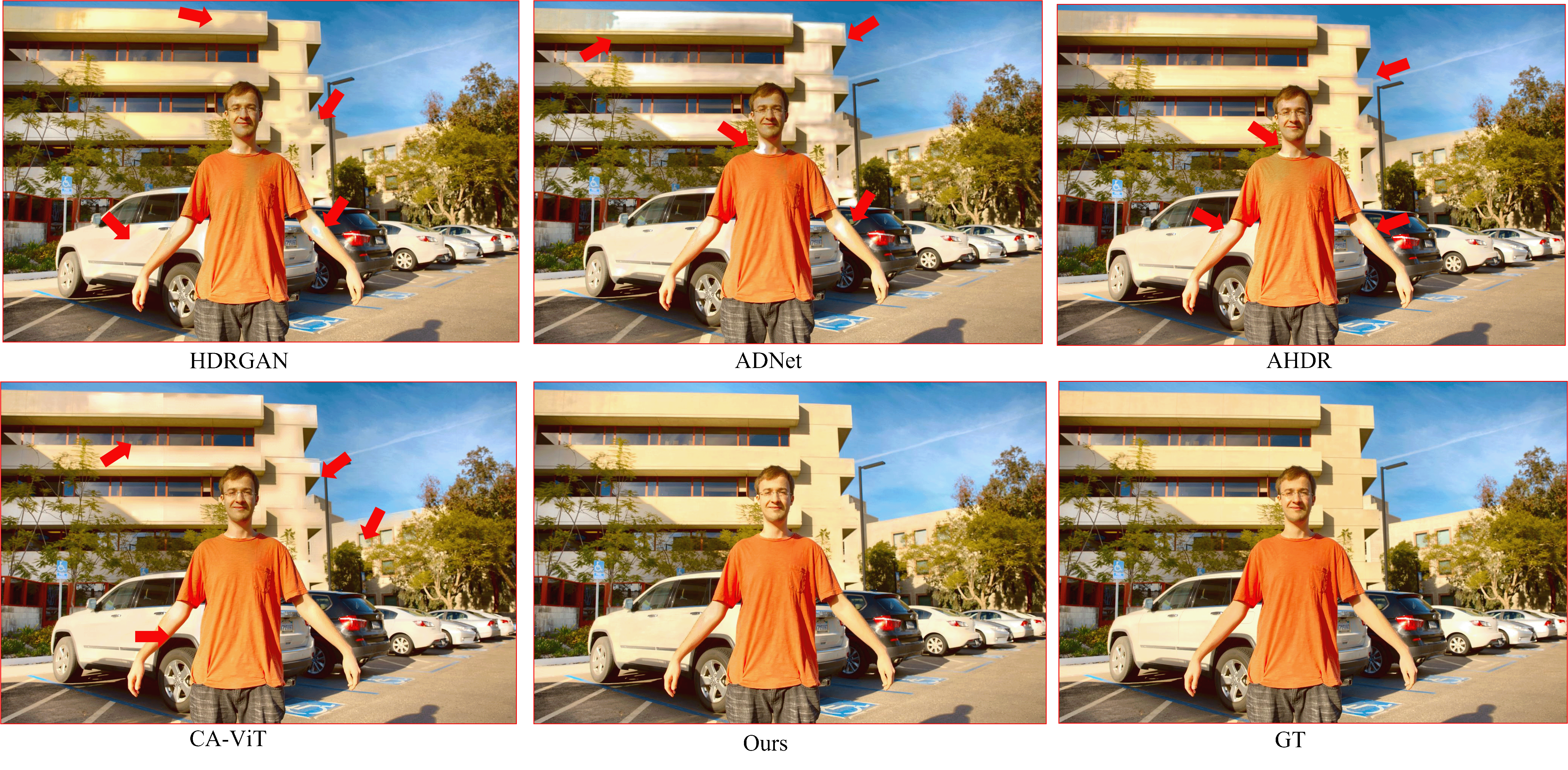}
\caption{Visual comparisons on the testing data from Kalantari \etal\cite{Kalantari2017Deep}. We compare the zoomed-in local areas of the HDR images estimated by our methods and the compared methods. Our model is capable of generating HDR images of exceptional quality, with results that closely align with how humans perceive visual information.}
\label{compare_full}
\vspace{-0.4cm}
\end{figure*}

\subsection{Experiment Settings}
\noindent\textbf{Datasets.} 
We conduct experiments on Kalantari's dataset \cite{Kalantari2017Deep}, which includes 74 samples for training and 15 samples for testing.
For each sample, three different LDR images are captured on a dynamic scene.
Following Yan \etal \cite{yan2019attention}, we choose three datasets for testing.
Specifically, we conduct the quantitative and qualitative evaluation on the 15 testing scenes in the Kalantari \etal’s dataset \cite{Kalantari2017Deep} with the provided ground truths. 
To further validate the generalizability of the model, we also use Sen \etal’s dataset \cite{Sen2012} and Tursun \etal’s dataset \cite{Tursun2016data} only for qualitative evaluation, which does not contain ground truths. 

\noindent\textbf{Evaluation Metrics.}
Five objective measures are employed to perform quantitative comparison: PSNR-$\mu$, SSIM-$\mu$, PSNR-L and SSIM-L, \xblue{and HDR-VDP-2 \cite{Mantiuk2011HDR} (calculated in the linear HDR domain)}, where $\mu$ and L means the metrics are calculated in the tonemapped domain and the linear domain, respectively.
We additionally computed two perceptual metrics, FID \cite{heusel2017gans} and LPIPS \cite{zhang2018unreasonable}. Due to the domain differences between HDR images and natural images, tonemapping was applied to both the generated results and ground truth (GT) to calculate the perceptual metrics. This allows for a more accurate evaluation of the quality of the generated images.

\noindent\textbf{Implementation Details.} We trained the model for 2,000,000 iterations. At each training iteration of diffusion models, we initially sampled 8 images from the training set and randomly cropped 8 patches of size 128$\times$128 from each, resulting in mini-batches of size 64. We use Adam optimizer \cite{Kingma2014Adam} with $\beta_1$ = 0.9, $\beta_2$= 0.999, and a fixed learning rate of $2e-5$. An exponential moving average weighing 0.999 was applied during parameter updates. All experiments are implemented based on the PyTorch~\cite{NEURIPS2019_9015} and carried out on NVIDIA A100 GPUs. Further specifications on the model
configurations are provided in Table.~\ref{npconfig}.

\subsection{Ghosting Reduction in HDR Imaging}

\xblue{To better illustrate the gradual effects of our method, we present intermediate results from the denoising process. During training, we adopt $T=1000$, resulting in 1000 intermediate results with adjacent ones having visually similar contents. We perform inference via the proposed SWNE method based on DDIM and present the intermediate results. As shown in Fig. \ref{inter result1} (a), we demonstrate the intermediate results at each step inferred by SWNE when $T=5$, which reveals how our method progressively denoises generating a high-quality HDR image from an input noise. Considering that our method predicts $x_0$ directly from the estimated noise at each denoising step and optimize it with an image space loss, we additionally present the results of directly predicting $x_0$ from each step's intermediate result, as shown in Fig. \ref{inter result1} (b). It can be observed that the quality of the predicted $x_0$ improves gradually as $t$ approaches 0, with less noise and finer details.}

\xblue{Furthermore, through the intermediate results, we conduct additional results to verify the effectiveness of DDPM for ghosting reduction in HDR imaging. We present the results of directly predicting $x_0$ from each intermediate step when $T=5$ using DDIM and SWNE. As depicted in Fig. \ref{inter result2}, the earlier results ($x_3,x_4$) have noticeable noise and ghosting artifacts. For example, ghosting artifacts can be observed in the arm region in Fig. \ref{inter result2} (a) and window sill in Fig. \ref{inter result2} (b), similar to AHDR \cite{yan2019attention}. With the denoising proceeds (from $x_4$ to $x_0$), while noise is reduced and details enhanced, the ghosting artifacts are progressively alleviated. We consider that this phenomenon can be attributed to the DDPM-based model having favorable content generation capabilities, thus it can remove ghosting and generate more perceptually pleasing HDR reconstructions compared to regression-based models.}

\subsection{Comparison with State-of-the-art Methods}
We evaluate the proposed method and compare it with state-of-the-art methods. Speciﬁcally, we compare the proposed method with a patch-based method \cite{Tursun2016data}, the ﬂow-based approach with DNN merger \cite{Kalantari2017Deep}, six CNN-based methods: DeepHDR \cite{Wu_2018_ECCV}, AHDR \cite{yan2019attention}, HDR-GAN \cite{niu2021hdr}, NHDRR \cite{yan2020deep}, ADNet \cite{liu2021adnet} and APNT \cite{chen2022attention}, and two ViT-based methods \cite{liu2022ghost,song2022selective}. The same training dataset and setting are used for deep learning methods.

\begin{figure}[tb]
\centering
\includegraphics[width=1\linewidth]{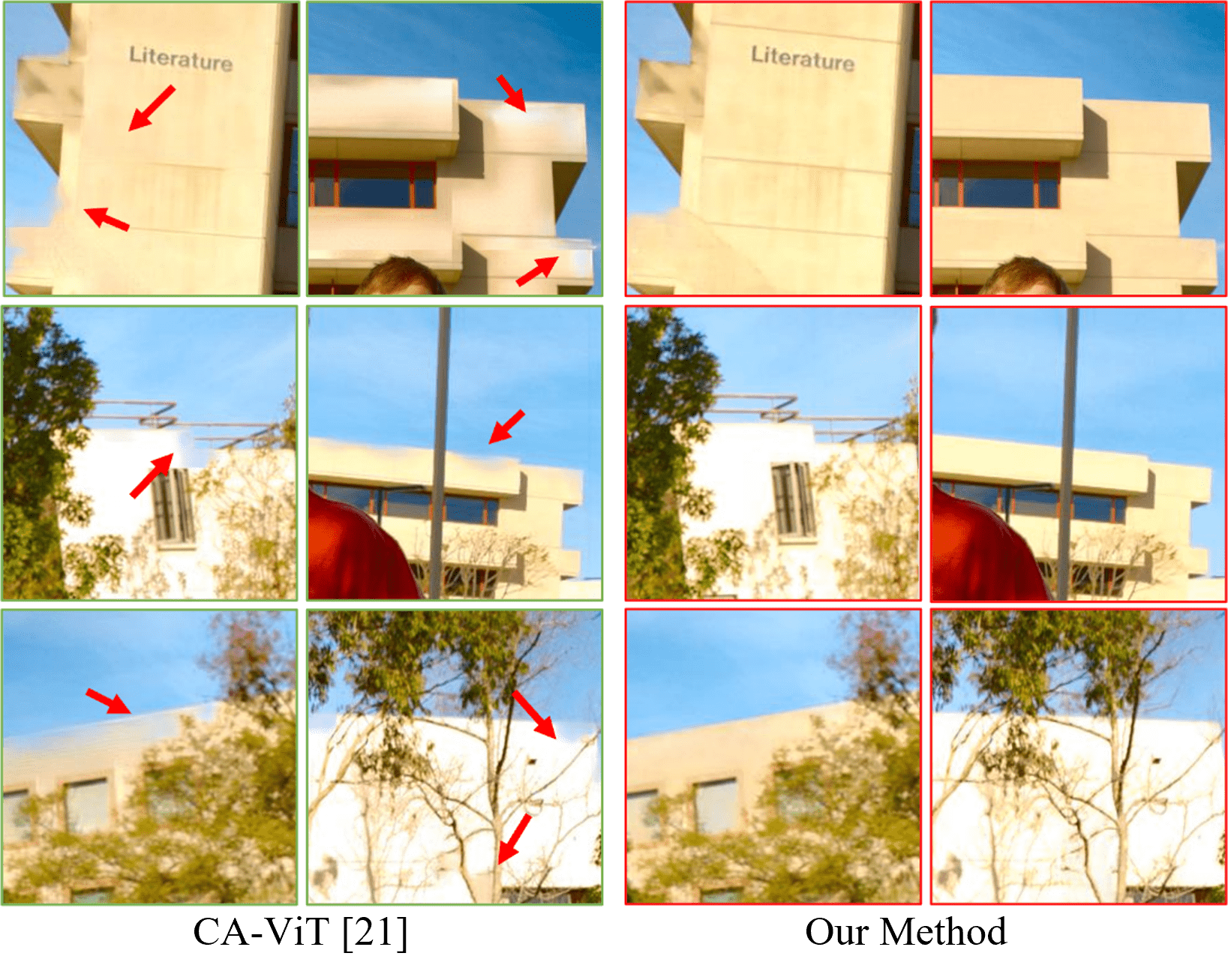}
\caption{Additional quality results comparisons with \cite{liu2022ghost}.}
\label{cavit}
\vspace{-0.4cm}
\end{figure}

\noindent\textbf{Results on Kalantari \etal’s Dataset.}
We compare our method with several state-of-the-art methods on the testing data of \cite{Kalantari2017Deep}, which contains some challenging samples with saturated background and foreground motions. For each test image, We set the step size $T$ as 25 to obtain the sampled results. All the quantitative results are averaged over 15 testing images. Note that in order to calculate FID and LPISP, we use publicly available models from related works. Official pre-trained models are provided by HDRGAN \cite{niu2021hdr}, and AHDR \cite{yan2019attention}, while ADNet \cite{liu2021adnet}, CA-ViT \cite{liu2022ghost}, and \cite{Kalantari2017Deep} were retrained using open source code.
Table \ref{compare} shows that our model outperforms previous models comprehensively in terms of perception-based metrics.

\subsubsection{Full images} We evaluate our method on the testing dataset, which contains some challenging samples with saturated background and foreground motions, and compare it with several state-of-the-art methods in Fig.~\ref{compare3}. Kalantari \etal \cite{Kalantari2017Deep} has better details than the CNN-based methods, but it can produce severe ghosting when the optical flow is not accurately estimated. The limitation of suppressing motion regions in AHDR\cite{yan2019attention} leads to the ghosting artifacts and the downsampling of NHDRR\cite{yan2020deep} results in damage details. HDR-GAN\cite{niu2021hdr}, ADNet\cite{liu2021adnet}, and APNT\cite{chen2022attention} all cannot completely handle the ghosting artifacts and also cannot restore the details in the saturated area. 
Furthermore, we conduct additional comparisons with \cite{liu2022ghost}. As shown in Fig. \ref{cavit}, our method possesses more image details and better handles the edges of buildings. Besides, due to patch-based sampling, \cite{liu2022ghost} may produce noticeable blocky ghosting (Fig. \ref{cavit} upper right corner). In contrast, our sliding window noise estimation does not introduce blocky ghosting and can generate HDR images that align with human perception.

\subsubsection{Challenging regions} The DDPM-based method has lower numerical results compared to the CNN-based baseline, which has been discussed in previous works \cite{saharia2022image}. In this paper, our method focuses on handling motion and saturation in LDR images, which is challenging. PSNR or SSIM cannot correlate well with human perception, thus we conducted additional human evaluation and reported distortion metrics for overexposed and moving areas as well as perception-based metrics for the full image compared to SOTA methods in Tables \ref{human study} and \ref{motion}.

\subsubsection{Results on Dataset with Cross-domain} 
\begin{figure*}[h]
\centering
\includegraphics[width=1\linewidth]{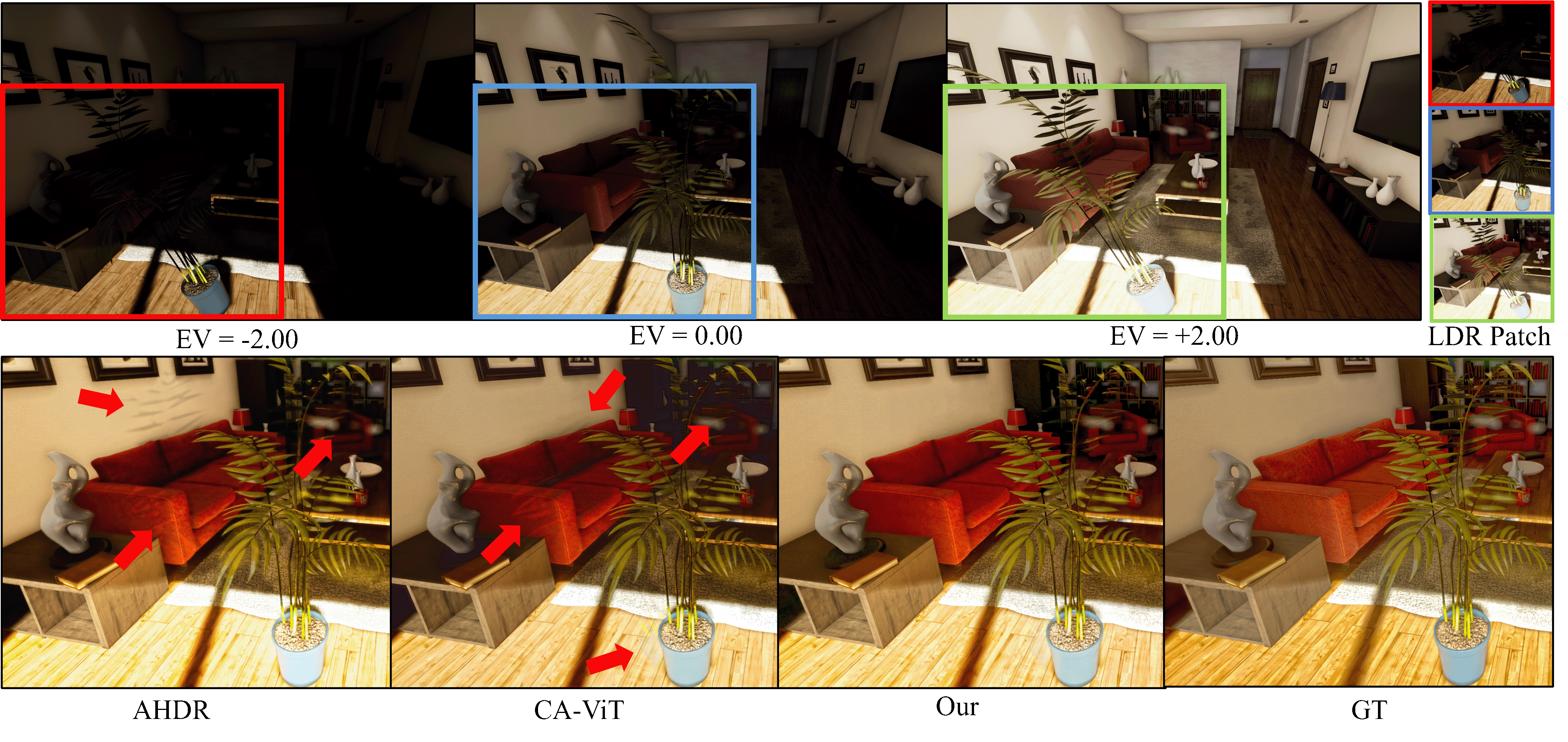}
\caption{Generalization experiment visualization comparison. After training on Kalantari \etal’s Dataset \cite{Kalantari2017Deep}, the model was directly evaluated on Hu \etal’s Dataset \cite{hu2020sensor} without any fine-tuning. It can be seen that our method has almost no ghosting and can recover more detailed information on the new dataset, even without fine-tuning.}
\label{compare_cross}
\end{figure*}

To further validate the generalization ability of our proposed model, we evaluated the model trained on Kalantari \etal’s Dataset \cite{Kalantari2017Deep} by testing it on Hu \etal’s Dataset \cite{hu2020sensor}. Since DDPM does not directly fit the target image but instead trains through perturbing data, it exhibits better generalization performance compared to CNN-based methods or Transformer-based methods. We compared our method (Table. \ref{cross test}) with the representative CNN-based model AHDR \cite{yan2019attention} and Transformer-based model CA-ViT \cite{liu2022ghost}. As shown in Fig. \ref{compare_cross}, we marked positions with noticeable ghosting artifacts and loss of details using red arrows. It can be seen that AHDR \cite{yan2019attention} have significant ghosting and loss of dark details, while CA-ViT \cite{liu2022ghost} alleviates ghosting to some extent but it is still noticeable, and similarly suffers from lost dark details. In contrast, our method has almost invisible ghosting and better detail recovery. Combined with results on images without ground truth, it can be seen that our method has better generalization performance.

\begin{table}[t]
\caption{The results of the challenging regions.}
\label{motion}
\scalebox{0.8}{
\begin{tabular}{c|cc|cc|cc}
\hline
       & \multicolumn{2}{c|}{Perception Metrics} & \multicolumn{2}{c|}{Miton+Oversaturation} & \multicolumn{2}{c}{Oversaturation} \\ \hline
Method & NIQE~$\downarrow$               & AHIQ~$\uparrow$               & PSNR-$\mu$~$\uparrow$          & PSNR-L~$\uparrow$              & PSNR-$\mu$       & PSNR-L~$\uparrow$          \\ \hline
ADNet  & 3.00               & 46,38              & \underline{35.47}               & 28.65               & 32.64            & 22.14               \\
AHDR   & 3.05               & 46.83              & 34.90               & 28.23               & 32.58            & 22.07               \\
HDRGAN & \underline{2.96}               & \underline{47.20}              & 34.62               & 27.84               & 32.72            & 22.07               \\
CA-ViT & 2.98               & 46.56              & 35.43               & \underline{29.21}               & \underline{33.20}            & \underline{22.62}  
   \\
Our    & \textbf{2.93}      & \textbf{47.55}     & \textbf{35.89}               &     \textbf{29.32}           & \textbf{33.25}            & \textbf{22.73}               \\ \hline
\end{tabular}
}
\end{table}

\begin{table}[t]
\caption{Cross-domain testing on Hu's dataset.}
\label{cross test}
\scalebox{0.9}{
\begin{tabular}{c|ccccc}
\hline
Method & PSNR-L~$\uparrow$ & PSNR-$\mu$~$\uparrow$ & SSIM-L~$\uparrow$ & SSIM-$\mu$~$\uparrow$ & LPIPS~$\downarrow$ \\ \hline
AHDR   & 24.80    & 20.90    & 92.99    & 81.99    & 0.0467     \\
CA-ViT & 27.08     & 21.65    & 93.02     & 84.00     & 0.0386      \\
Our    & \textbf{29.23}    & \textbf{22.25}    & \textbf{94.13}    & \textbf{84.09}    & \textbf{0.0302}     \\ \hline
\end{tabular}
}
\end{table}

\begin{figure*}[tb]
\centering
\includegraphics[width=1\linewidth]{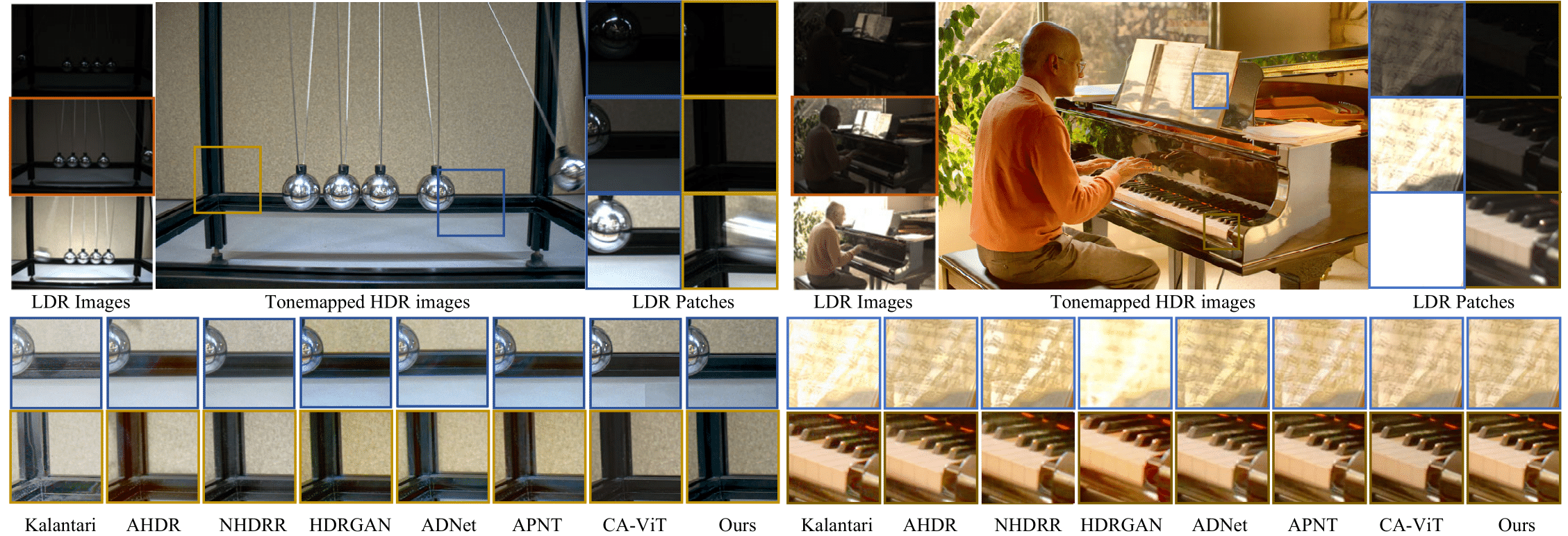}
\footnotesize (a) Example from Sen \etal's dataset \cite{Sen2012} \qquad\qquad\qquad\qquad\qquad\qquad\qquad\qquad (b) Example from Tursen \etal's dataset \cite{Tursun2016data}
\includegraphics[width=1\linewidth]{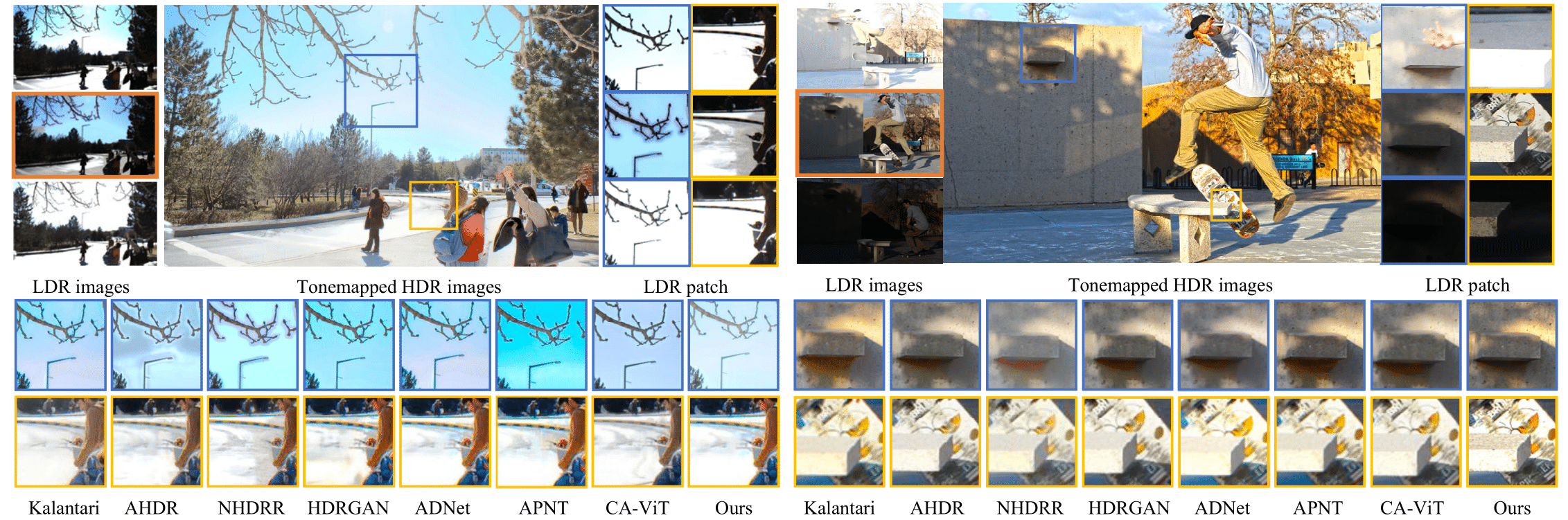}
\footnotesize (c) Example from Sen \etal's dataset \cite{Sen2012} \qquad\qquad\qquad\qquad\qquad\qquad\qquad\qquad (d) Example from Tursen \etal's dataset \cite{Tursun2016data}
\caption{Visual comparisons on the datasets without ground truth.}
\label{comparewogt}
\end{figure*}

\noindent\textbf{Results on Dataset without Ground Truth.}
\xblue{To verify the generalization ability of the proposed method for HDR imaging, we also evaluate the performance on Sen \etal \cite{sen2012robust}’s and Tursun \etal \cite{tursun2016objective}’s datasets which do not provide the ground truth. The results are shown in Fig.~\ref{comparewogt}. According to Fig. \ref{comparewogt} (a): In the orange block, ghosting artifacts are present in all methods except for HDRGAN \cite{niu2021hdr}, CA-Vit \cite{liu2022ghost}, and our proposed method. However, \cite{liu2022ghost} suffers from block artifacts due to its patch-based approach. In the blue block, inaccurate color details exist in all methods except for our proposed method and CA-ViT \cite{liu2022ghost}. In Fig.~\ref{comparewogt} (b), some methods, such as Kalantari \cite{Kalantari2017Deep} and HDR-GAN \cite{niu2021hdr}, are unable to process saturated areas. NHDRR \cite{yan2020deep} and APNT \cite{chen2022attention} partially eliminate saturated areas, while AHDR \cite{yan2019attention} and ADNet \cite{liu2021adnet} lose some details in the piano. 
In Fig. \ref{comparewogt} (c), some methods exhibit obvious ghosting artifacts within the orange block, including AHDR \cite{yan2019attention}, HDRGAN \cite{niu2021hdr}, and CA-ViT \cite{liu2022ghost}. While Kalantari's method \cite{Kalantari2017Deep} does not have ghosting, it suffers from noticeable color distortion. In Fig. \ref{comparewogt} (d), although no method has conspicuous ghosting, our approach produces sharper image details compared to previous techniques.}

\subsection{Ablation Study}
\label{ablation sec}

\begin{figure}[tb]
\centering
\includegraphics[width=1\linewidth]{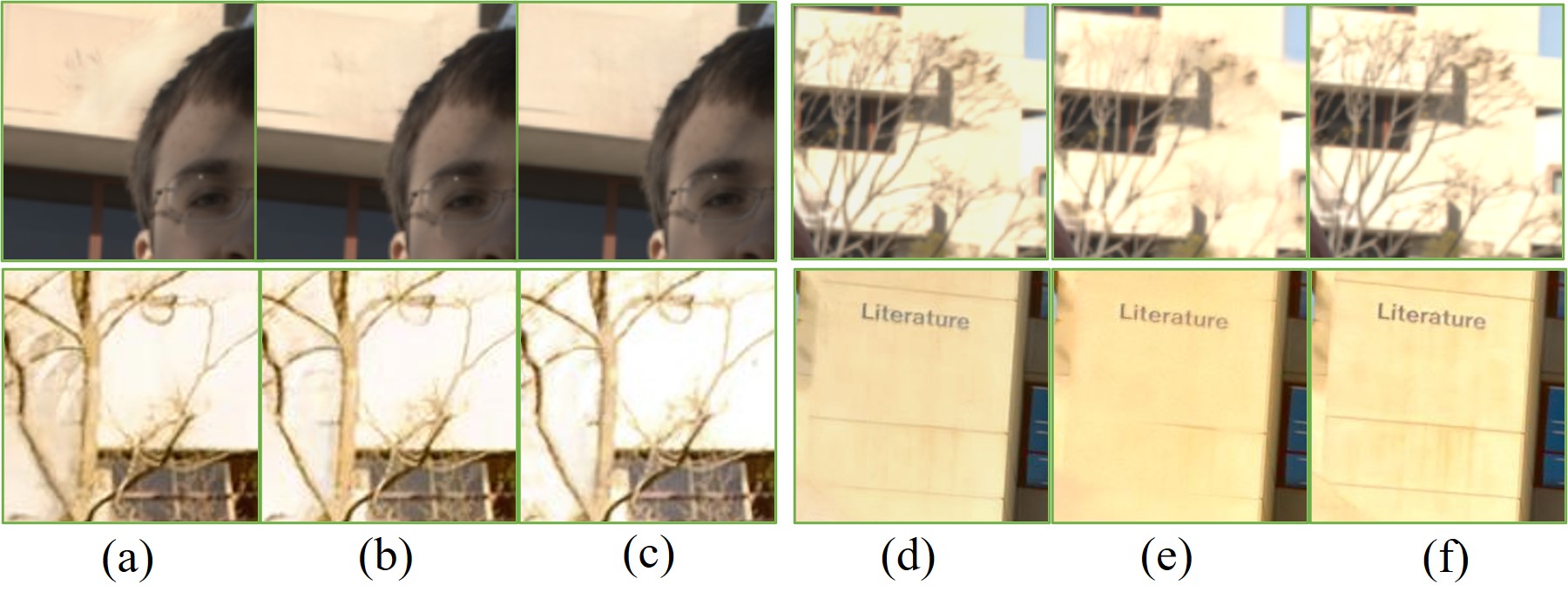}
\caption{Qualitative results of our ablation study on the proposed components.}
\label{ablationpng}
\end{figure}

We perform ablation learning on different components to validate their effectiveness.

\noindent\textbf{Study on FCG.} Considering the domain differences between noisy image $x_t$ and LDR images, as well as the motion variations among different LDR images, our FCG module comprises two aspects: feature alignment and feature embedding. We have introduced two variants for this purpose: (1) directly concatenating different LDR images with $x_t$, and (2) removing the DFA layer and concatenating aligned features with $x_t$, which has undergone one convolutional layer. As shown in Fig. \ref{ablationpng}, we observed that Variant 1 (Fig. \ref{ablationpng} (a)) led to the confusion or loss of information between different LDR images, resulting in noticeable ghosting artifacts in the motion regions. By deploying the AM module to align features, Variant 2 (Fig.~\ref{ablationpng} (b)) enables more accurate extraction of features preserved in different LDR images, thereby avoiding noticeable ghosting artifacts. However, blurring still exists in the motion regions. Adding the DFA layer results (Fig. \ref{ablationpng} (c)) in a cleaner effect in the motion regions.

\noindent\textbf{Study on SWNE.} 
SWNE controls the size of the receptive field during sampling to focus on local contextual information, resulting in better-quality generated results. Since SWNE is a sampling method that does not involve training, we evaluated the impact of patch size on model performance using the same weights and six different combinations of $P$ and $r$. As shown in Table. \ref{ablation1}, the model achieved the best overall performance when the patch size was 512. Increasing or decreasing $P$ had a certain impact on the model's overall performance. This could be because oversized patches caused the model to generate errors in the exposed area by utilizing distant information, while undersized patches led to reduced performance due to the limited receptive field that couldn't fully utilize the surrounding pixel information.

\begin{figure}[t]
\centering
\includegraphics[width=1\linewidth]{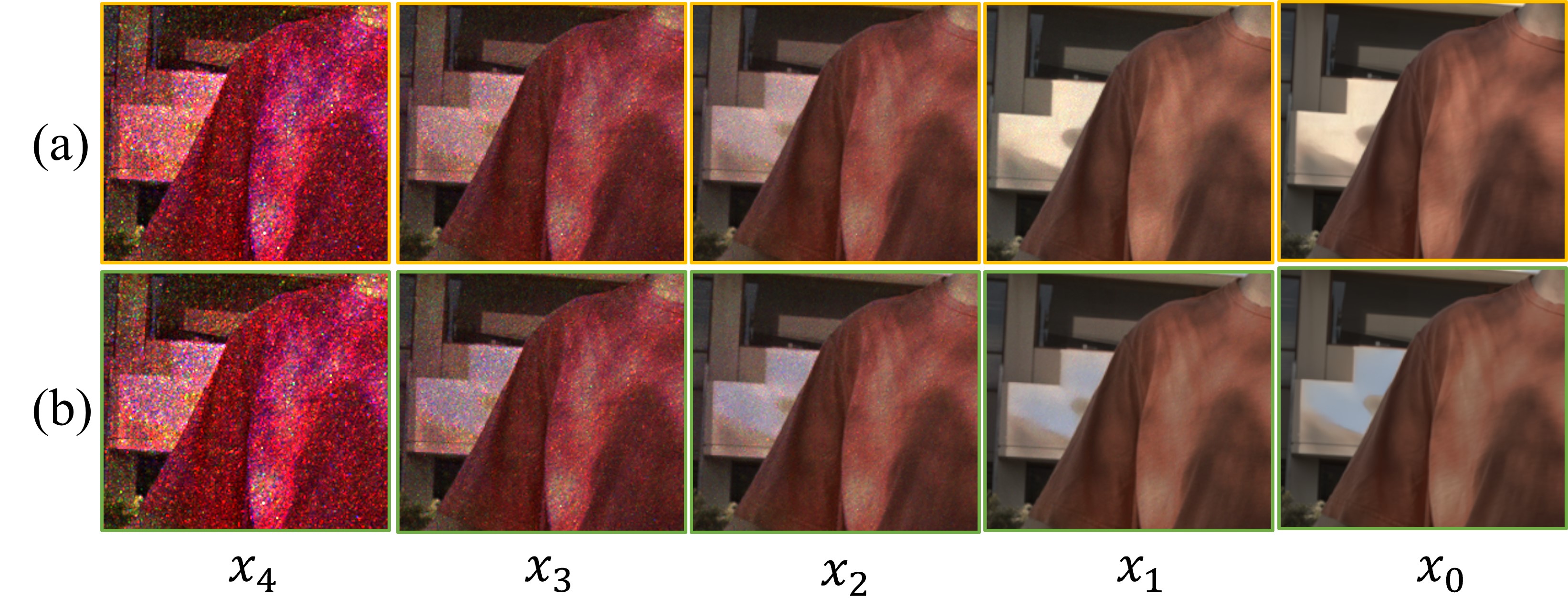}
\caption{Ablation studies on the impact of SWNE during inference. The figure shows denoising results from the intermediate step of directly predicting $x_0$ from $x_t$ using (a) SWNE with patch size 512 and stride 128 during inference, and (b) whole image inference without SWNE.}
\label{swne_vis}
\end{figure}

\xblue{In addition, we use DDIM and set $T=5$ to present the intermediate results of directly predicting $x_0$ at each denoising step. As shown in Fig. \ref{swne_vis}, $x_4$ to $x_0$ denote the intermediate results at different steps, with reduced noise and increased details as denoising proceeds. Fig. \ref{swne_vis} (a) shows results using $512\times 512$ patch-based SWNE inference, while Fig. \ref{swne_vis} (b) uses full-image inference without SWNE. It can be observed that in the early phases of the denoising step (\eg,$x_4$), the variants have very similar outputs, but as denoising continues, full-image inference utilizes erroneous context to reconstruct saturated regions, leading to unrealistic results inconsistent with human perception. While leveraging broader context can help reconstruct regions lacking reliable local information to some extent, the semantics of distant locations often differ from the local patch. In contrast, the SWNE approach (Fig. \ref{swne_vis} (a)) encourages the model to fully utilize the learned patch statistics and focus on local contextual information, thus avoiding semantic confusion and producing more plausible HDR reconstructions.}

\noindent\textbf{Study on Loss Function.} We perform an ablation study using various combinations of loss functions in Table \ref{ablation2}, and our findings are detailed in Fig. \ref{ablationpng}. Our results indicate that the exclusive use of $\mathcal{L}_{noise}$ (Fig.~\ref{ablationpng} (d)) can generate image details that align with human perception. However, it results in a noticeable color shift. Conversely, employing only $\mathcal{L}_{image}$ results (Fig.~\ref{ablationpng} (e)) in a loss of details and creates blurry images. This outcome is unsurprising given the tendencies of image-based $MSE$ regression approaches to penalize any produced high-frequency details that are not perfectly matched with the target image. As perfect alignment is almost impossible when synthesizing high-frequency details such as the identical wall texture depicted in Fig.~\ref{ablationpng}, the resulting images are further blurred. This issue has been observed in previous CNN-based methods. Therefore, a better option than relying on any single loss function alone is to use a combination of loss functions (Fig.~\ref{ablationpng} (f)) to strike a balance between image details and color information.

\begin{figure*}[t]
\centering
\includegraphics[width=1\linewidth]{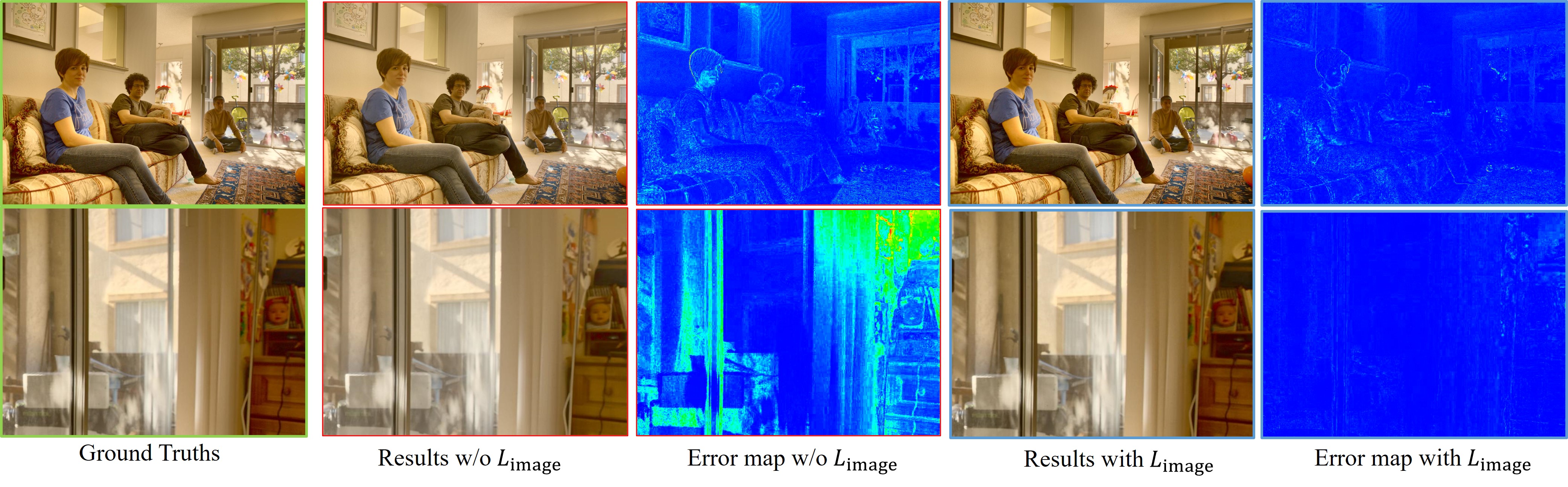}
\caption{Ablation studies on the impact of $\mathcal{L}_{image}$. The difference in reconstructed content structure with or without $\mathcal{L}{image}$ is barely noticeable. However, using $\mathcal{L}{image}$ helps avoid color distortion in the reconstructed contents, as can be observed from the error maps.}
\label{loss visul}
\end{figure*}

\noindent \xblue{\textbf{Study on $\mathcal{L}_{image}$.~}}\xblue{ Since the proposed $\mathcal{L}_{image}$ can provide the constraints with pixel level, it effectively addresses color distortion issues in the DDPM-based model.
To verify this point, we conducted several ablation results as shown in Fig. \ref{loss visul}, which presents HDR reconstructions and error maps with and without $\mathcal{L}_{image}$. It can be observed that while both variants produce HDR images with content closely matching the ground truth, the result without $\mathcal{L}_{image}$ exhibits noticeable color distortion from the ground truth. In contrast, the HDR image reconstructed with $\mathcal{L}_{image}$ has colors closer to the real scene, as shown in the error map in Fig. \ref{loss visul}.}

\begin{table}[!tp]\small
\caption{Ablation studies on the Sliding-window Noise Estimation and loss functions.}
\centering
\small
\scalebox{0.75}{
\begin{tabular}{cc|cccccc}
\hline
$\mathcal{L}_{noise}$ & $\mathcal{L}_{image}$ & $r$            & $P$             & PSNR-$\mu$~$\uparrow$     & PSNR-L~$\uparrow$         & SSIM-L~$\uparrow$        & SSIM-$\mu$~$\uparrow$     \\ \hline
\checkmark              & \checkmark              & 16            & 64            & 42.62          & 41.15          & 98.80          & 99.05          \\
\checkmark              & \checkmark              & 32            & 128           & 43.71          & 41.61          & 98.85          & \textbf{99.14} \\
\checkmark              & \checkmark              & 64            & 256           & 43.92          & 41.71          & \textbf{98.88} & 99.13          \\
\checkmark               & \checkmark              & 128           & 512           & \textbf{44.11} & \textbf{41.73} & 98.85          & 99.11          \\
\checkmark              & \checkmark              & 256           & 1024          & 43.44          & 41.39          & 98.75          & 98.95          \\
\checkmark              & \checkmark              & \multicolumn{2}{c}{full size} & 43.29          & 41.31          & 98.73          & 98.94          \\ \hline
\checkmark              &                       & 128           & 512           & 36.88          & 38.12          & 94.02          & 95.60          \\
                      & \checkmark              & 128           & 512           & 43.64          & 41.59          & 98.83          & 99.01          \\
\checkmark              & \checkmark              & 128           & 512           & \textbf{44.11} & \textbf{41.73} & \textbf{98.85} & \textbf{99.11} \\ \hline
\end{tabular}
}
\label{ablation1}
\end{table}

\begin{table}[!t]\small
\caption{Ablation studies on the feature condition generator.}
\centering
\scalebox{0.9}{
\begin{tabular}{c|cccc}
\hline
      & PSNR-$\mu$~$\uparrow$ & PSNR-L~$\uparrow$ & PSNR-L~$\uparrow$ & SSIM-$\mu$~$\uparrow$ \\ \hline
Concat & 43.17    & 41.02    & 98.63    & 98.82    \\
Concat + AM    & 44.02    & 41.72    & 98.74    & 99.07    \\
AM + DFA   & \textbf{44.11}    & \textbf{41.73}    & \textbf{98.85}    & \textbf{99.11}    \\ \hline
\end{tabular}}
\label{ablation2}
\end{table}

\subsection{Human Study for Qualitative Evaluation}
\xblue{We conducted a perceptual study with human subjects to further evaluate the performance of the proposed HDR imaging framework. To obtain pairwise preference ratings comparing different HDR imaging methods on the Kalantari \cite{Kalantari2017Deep} dataset, we presented participants with side-by-side $512\times 512$ crops of images generated by each method. Specifically, we decoupled the 15 imaging results from each model into $15 \times 6=90$ patches of size 512$\times$512. Since HDR images are usually displayed after tone mapping, we visualized the images after applying MATLAB's tonemap operation to convert them to the tonemapped images. The images were displayed on a 27-inch LED monitor with $2560\times 1440$ resolution. To ensure suitable participants, we employed the Snellen visual acuity test and a color blindness assessment. A total of 30 graduate student participants (18 females, 12 males) aged 18-30 years met the criteria. To improve rating consistency, we included a training step before the rating task to familiarize participants with the distortion types present in the test. There were 90 independent image groups, each containing crops from the same scene location generated by different methods. Participants were asked to select the image with the better quality from side-by-side crops of size $512 \times 512$.}

\begin{table}[tbp]
  \centering
  
  \caption{ Quantitative comparison in the human study.
  }
\scalebox{0.9}{
    \begin{tabular}{ccccccc}
    \hline
          & Label & HU & HDRGAN
          & CA-ViT & AHDRNet & Ours  \\
    \hline
    Label & -     & 89.70& 75.50 & 70.29 & 80.56 & 59.09  \\
    HU    & 10.30 & -     & 26.14 & 21.36 & 32.24 & 14.22  \\
    HDRGAN & 24.50 & 73.86 & -     & 43.42 & 57.35 & 31.91  \\
    CA-ViT & 29.71 & 78.64 & 56.58 & -     & 63.65 & 37.91  \\
    AHDRNet & 19.44 & 67.76 & 42.65 & 36,35 & - & 25.85 \\
    Ours  & \textbf{40.95} & \textbf{85.78} & \textbf{68.09} & \textbf{62.09} & \textbf{74.15} & -      \\
    \hline
    \end{tabular}}
  \label{human study}
\end{table}

\xblue{Results in Tab. \ref{human study} show the average rater preference computed from $30\times 90=2700$ comparisons. Each value represents the percentage of times raters preferred the row over the column. As highlighted, these results indicate that our HDR imaging model outperformed the competing methods.}


\begin{table*}[t]
\begin{center}
\renewcommand{\arraystretch}{1.3}
\caption{Average running time for different methods on the testing set with size $1000 \times 1500$ under GPU or CPU environment.}
  \label{inftime}
  \scalebox{0.9}{
  \begin{tabular}{c|c|c|c|c|c|c|c|c|c}
  \hline
  Methods &Sen \cite{sen2012robust} & Hu \cite{hu2013hdr}& Kalantari \cite{Kalantari2017Deep} &  AHDR \cite{yan2019attention}&  HDRGAN \cite{niu2021hdr}&  CA-ViT \cite{liu2022ghost} &w/o DDIM&w DDIM \cite{DBLP:conf/iclr/SongME21}&w DPM-solver \cite{lu2022dpm}\\
  Environment &(CPU)&(CPU) &(CPU+GPU)&(GPU)&(GPU)&(GPU)&(GPU)&(GPU)&(GPU) \\
     \hline
     Times(s) & 61.81s & 79.77s & 29.14s & 0.30s & 0.29s& 5.34s& 293.48s& 7.53s& 0.59s \\
     \hline
\end{tabular}}
\end{center}
\end{table*}

\section{Discussion}
\subsection{Tonemapped Denoising Target}
\label{tonemap diss}
\begin{figure*}[h]
\centering
\includegraphics[width=1\linewidth]{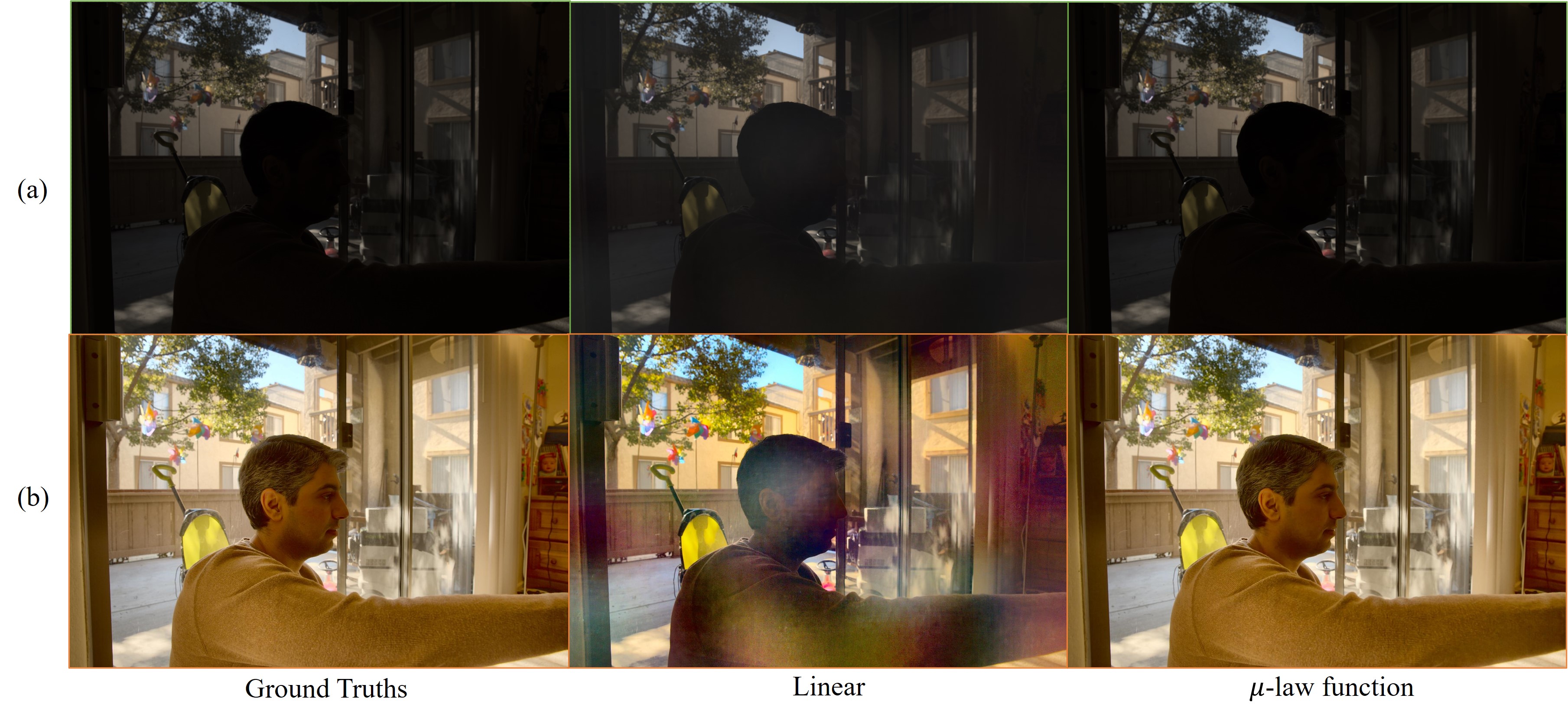}
\caption{Tonemapping boosts the pixel values in the dark regions, and thus, optimization in the tonemapped domain gives more emphasis to these darker pixels in comparison with the optimization in the linear domain.}
\label{mulawour}
\end{figure*}
\xblue{In previous work, \cite{Kalantari2017Deep} found that when a model produces HDR images in the linear HDR domain, the estimated images typically demonstrate discoloration, noise, and other artifacts. Therefore, they proposed training the model in a more uniform amplitude distribution domain to obtain better visual results. Given the estimated HDR image $I^{\hat{H}}$ and the ground truth HDR image $I^H$, this loss term is defined as:
\begin{equation}
\mathcal{L}=\left\|\mathcal{T}\left(I^H\right)-\mathcal{T}\left(I^{\hat{H}}\right)\right\|_1,
\label{tloss4}
\end{equation}
where $\mathcal{T}(\cdot)$ denotes the $\mu$-law function, and  $\mu=5000$. 
In our work, since traditional DDPM-based methods only learn a probability distribution of the added noise in each step, we cannot directly apply the loss function in Eq. \ref{tloss4}. Therefore, we devised a ddpm-specific approach that applies the $\mu$-law function to transform the HDR image before adding noise, in order to indirectly achieve a similar effect. In our initial experiments training directly on the linear HDR domain, results exhibited discoloration, noise, and artifacts in dark regions. As seen in Fig. \ref{mulawour}, while HDR domain results (Fig. \ref{mulawour} (a)) appear similar, tonemapped images (Fig. \ref{mulawour} (b)) reveal significant artifacts when trained on the linear HDR domain. The $\mu$-law function boosts the pixel values in the dark regions, and thus, optimization in the tonemapped domain gives more emphasis to these darker pixels in comparison with the optimization in the linear domain. It is worth noting that, as discussed in \cite{Kalantari2017Deep}, tonemapping does not directly and irreversibly change the fundamental problem, unless the tonemapping operator chosen is non-differentiable or irreversible. For example, Gamma encoding, defined as ${(I^H)}^{1 / \gamma}$ with $\gamma>1$, is perhaps the simplest way of tonemapping in image processing. However, since it is not differentiable around zero, it is not suitable for use in our system.}

\subsection{Analysis of Computational Budgets}
\xblue{As our method aims to explore the capabilities of DDPM for HDR imaging, we only adopted the common DDIM \cite{DBLP:conf/iclr/SongME21} algorithm to accelerate model inference in this work. Without any acceleration algorithms, the iterative sampling process of DDPM does require more time to generate the final HDR results compared to prior deep learning approaches. Thus, the standard DDPM paradigm presents challenges for practical deployment on lightweight devices. To address this issue, both academia and industry have initiated research into accelerating diffusion models along two main directions: (1) reducing the number of inference steps \cite{DBLP:conf/iclr/SongME21,lu2022dpm,meng2023distillation}; and (2) engineering optimizations \cite{chen2023speed,li2023snapfusion}. Since our method builds upon the same theoretical framework, these acceleration techniques can be seamlessly applied. With state-of-the-art strategies \cite{li2023snapfusion}, inference of billion-parameter DDPM models can be achieved on mobile devices within 2 seconds without compromising image quality.}

\xblue{We benchmarked the inference time under different strategies to provide quantitative comparisons (Tab. \ref{inftime}). Without any acceleration, our DDPM model takes 293 seconds to generate a 1500$\times$1000 HDR image using 1000 sampling steps. By adopting DDIM \cite{DBLP:conf/iclr/SongME21}, we can achieve comparable visual results with only 25 steps, reducing the inference time to 7.53 seconds. Using DPM Solver \cite{lu2022dpm} allows further acceleration, producing comparable quality with just 5 steps and 0.59 seconds. While slower than some existing techniques, this demonstrates the potential to significantly decrease running time while maintaining visual fidelity. Designing DDPM-based methods suitable for lightweight devices will be an important direction for our future work.}


\section{Conclusion}
In this paper, we change perspective and consider HDR imaging a conditional generative modeling task. We presented a new framework for HDR reconstruction from multi-exposed LDR images and focused on perceptual quality using a conditional diffusion model. By adopting the stochastic iterative denoising process, our method can produce reliable information for HDR images when LDR images contain large object motions. Extensive experiments on benchmark datasets for HDR imaging demonstrate that the proposed method achieves state-of-the-art performances and well generalizations to real-world images. Due to the iterative generation of DDPM, its inference speed is relatively slow. Additionally, although its generated output exhibits better visual effects compared to previous methods, there is still room for improvement in terms of distortion-based metrics. This indicates a promising research direction for future works. \\
\section*{Acknowledgment}
This work is supported by NSFC of China 62301432, 6230624, Natural Science Basic Research Program of Shaanxi No. 2023-JC-QN-0685, QCYRCXM-2023-057, the Fundamental Research Funds for the Central Universities No. D5000220444.

\bibliographystyle{IEEEtran}
\bibliography{egbib.bib}

\vfill

\end{document}